\documentclass[letterpaper]{article} 
\usepackage{aaai24}  
\usepackage{times}  
\usepackage{helvet}  
\usepackage{courier}  
\usepackage[hyphens]{url}  
\usepackage{graphicx} 
\usepackage{natbib}  
\usepackage{caption} 
\usepackage{algorithm}
\usepackage{algorithmic}

\usepackage{amsmath}
\usepackage{amssymb}
\usepackage{multirow}
\usepackage{booktabs}
\usepackage{makecell}

\usepackage{xcolor}

\usepackage{pifont}
\newcommand{\figref}[1]{Fig.~\ref{#1}}

\newcommand{\tableref}[1]{Table~\ref{#1}}

\newcommand{\yu}{\textcolor[rgb]{0., 0., 0}}
\newcommand{\yur}{\textcolor[rgb]{0,0,0}}

\newcommand{\kk}{\textcolor[rgb]{.0,.0,.0}}
\newcommand{\ryn}{\textcolor[rgb]{0,0,0.0}}

\newcommand{\ke}{\textcolor[rgb]{0.0,0.0,0.0}}

\newcommand{\rep}{\textcolor[rgb]{0, 0.0, 0}}
\newcommand{\rynn}{\textcolor[rgb]{0,0,0}}
\newcommand{\rynnq}{\textcolor[rgb]{0,0,0}}

\makeatletter
\DeclareRobustCommand\onedot{\futurelet\@let@token\@onedot}
\def\@onedot{\ifx\@let@token.\else.\null\fi\xspace}
\def\eg{\emph{e.g}\onedot} 
\def\ie{\emph{i.e}\onedot}

\makeatother

\def\ie{\textit{i.e.}}
\def\eg{\textit{e.g.}}

\def\et{\textit{et al.}}

\usepackage{newfloat}
\usepackage{listings}
\DeclareCaptionStyle{ruled}{labelfont=normalfont,labelsep=colon,strut=off} %
\floatstyle{ruled}
\newfloat{listing}{tb}{lst}{}
\floatname{listing}{Listing}
\title{Recasting Regional Lighting for Shadow Removal}
\author{
    Yuhao Liu, 
    Zhanghan Ke, 
    Ke Xu\thanks{Joint corresponding authors. Rynson Lau leads this project.}, 
    Fang Liu, 
    Zhenwei Wang, 
    Rynson W.H. Lau$^*$
}
\affiliations{
    Department of Computer Science, City University of Hong Kong
}

\usepackage{bibentry}

\begin{document}

\maketitle

\begin{abstract}
Removing shadows requires an understanding of both lighting conditions and \ryn{object textures} in a scene.
Existing methods typically learn pixel-level color mappings between shadow and non-shadow images, in which the joint modeling of lighting and object \ryn{textures} is implicit and \ryn{inadequate}.
We observe that in a shadow region, the degradation degree of object textures depends on the local illumination, while \rynn{simply enhancing the local illumination} cannot fully recover the attenuated textures.
\ryn{Based on this observation,} \yu{we propose 
to condition the restoration of attenuated textures on the corrected local lighting in the shadow region.} \yu{Specifically, \kk{We}} first \kk{design} a shadow-aware decomposition network to estimate the illumination and reflectance layers of shadow regions explicitly. \kk{We then propose} a novel bilateral correction network \rynn{to recast} the lighting of shadow regions in the illumination layer via a novel local lighting correction module, and \rynn{to restore} the textures conditioned on the corrected illumination layer via a novel illumination-guided texture restoration module. 
\ryn{We further annotate pixel-wise shadow masks for the SRD dataset, which originally contains only image pairs.}
Experiments on three benchmarks show that 
our method outperforms \ryn{existing SOTA} shadow removal methods. 
\end{abstract}

\section{Introduction}
\label{sec:introduction}

Shadows manifest on surfaces where light is partially or entirely blocked, resulting in image areas with reduced intensity, darker colors, and diminished textures. These shadows can create recognition ambiguities in existing visual models, such as text recognition~\cite{brown2006geometric}, remote traffic monitoring~\cite{zhang2020shadow}, and object localization~\cite{mei2021camouflaged,liu2023referring}. Consequently, the study of shadow removal becomes crucial.

There are a \ryn{number of} shadow removal methods proposed.
Previous non-deep learning-based methods~\cite{finlayson2002removing,guo2012paired,gryka2015learning,finlayson2009entropy,finlayson2001,finlayson2005removal,yang2012shadow,zhang2015shadow} typically \ryn{use} hand-crafted priors and/or leverage user interactions to remove shadows, which often fail in complex real-world scenes~\cite{khan2015automatic}.

\begin{figure}[t!]
\begin{center}
\begin{tabular}{c@{\hspace{0.8mm}}c@{\hspace{0.8mm}}c@{\hspace{0.8mm}}c@{\hspace{0.8mm}}c}
\includegraphics[width=0.23\linewidth,height=0.17\linewidth]{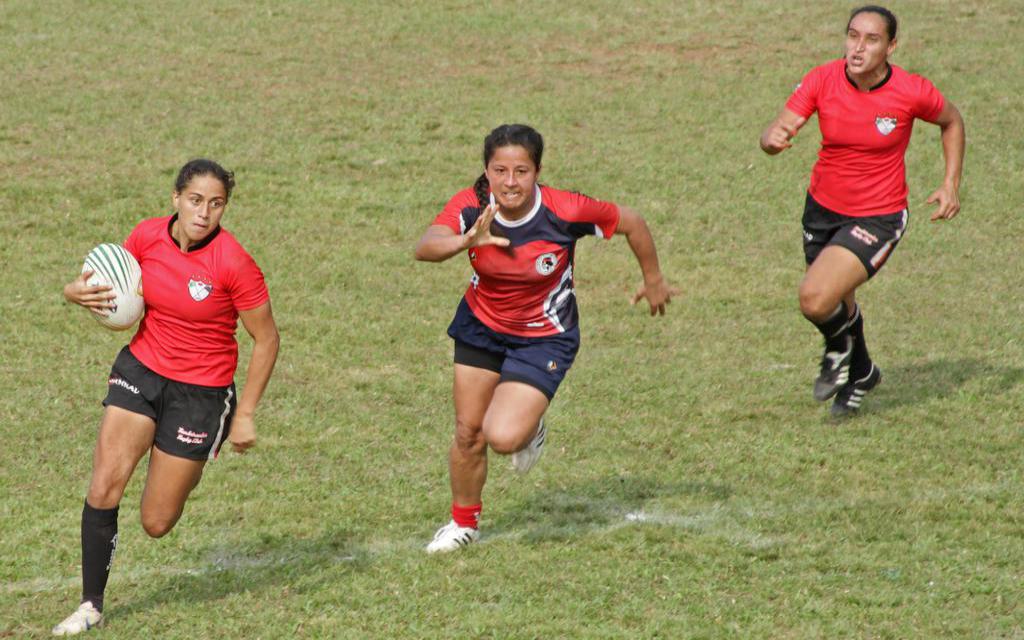}&
\includegraphics[width=0.23\linewidth,height=0.17\linewidth]{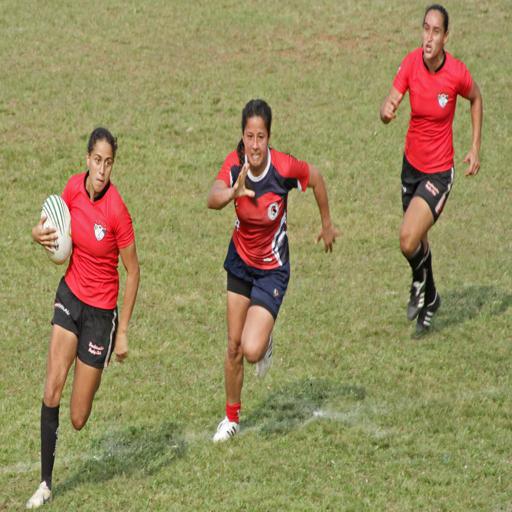}&
\includegraphics[width=0.23\linewidth,height=0.17\linewidth]{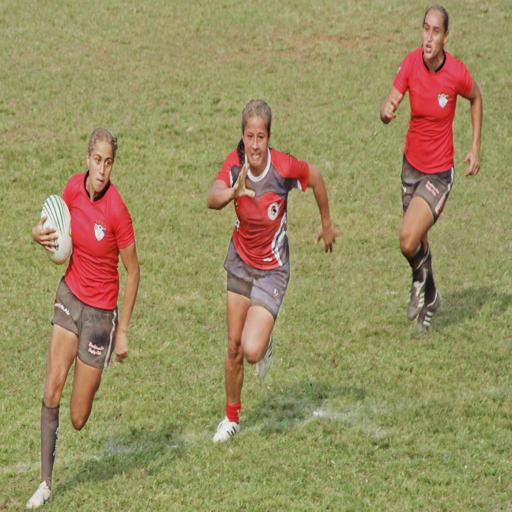}&
\includegraphics[width=0.23\linewidth,height=0.17\linewidth]{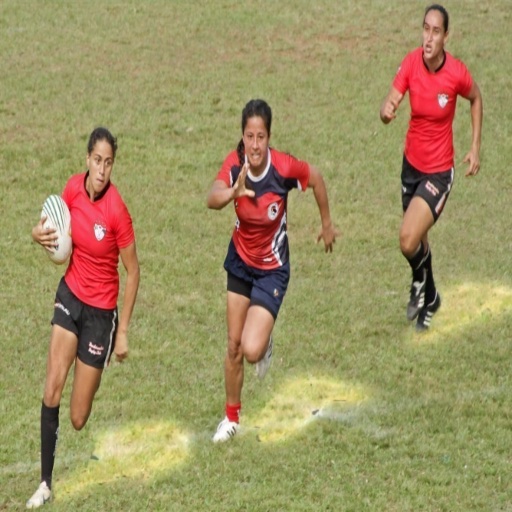}& 

\\

\fontsize{8pt}{\baselineskip}\selectfont (a) Shadow &
\fontsize{8pt}{\baselineskip}\selectfont (b) AEF &
\fontsize{8pt}{\baselineskip}\selectfont  (c) DCGAN &
\fontsize{8pt}{\baselineskip}\selectfont  (d) BMNet &

\\

\includegraphics[width=0.23\linewidth,height=0.17\linewidth]{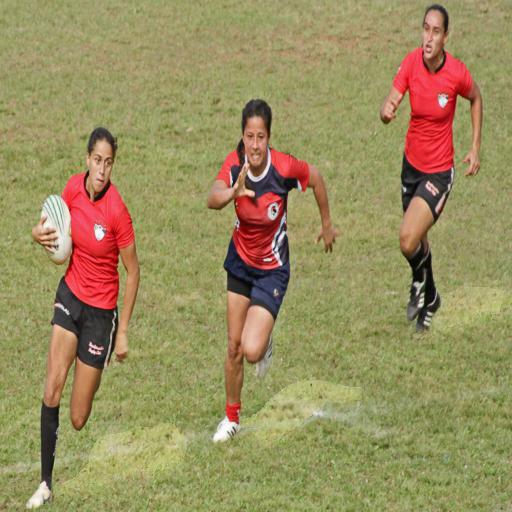}& 
\includegraphics[width=0.23\linewidth,height=0.17\linewidth]{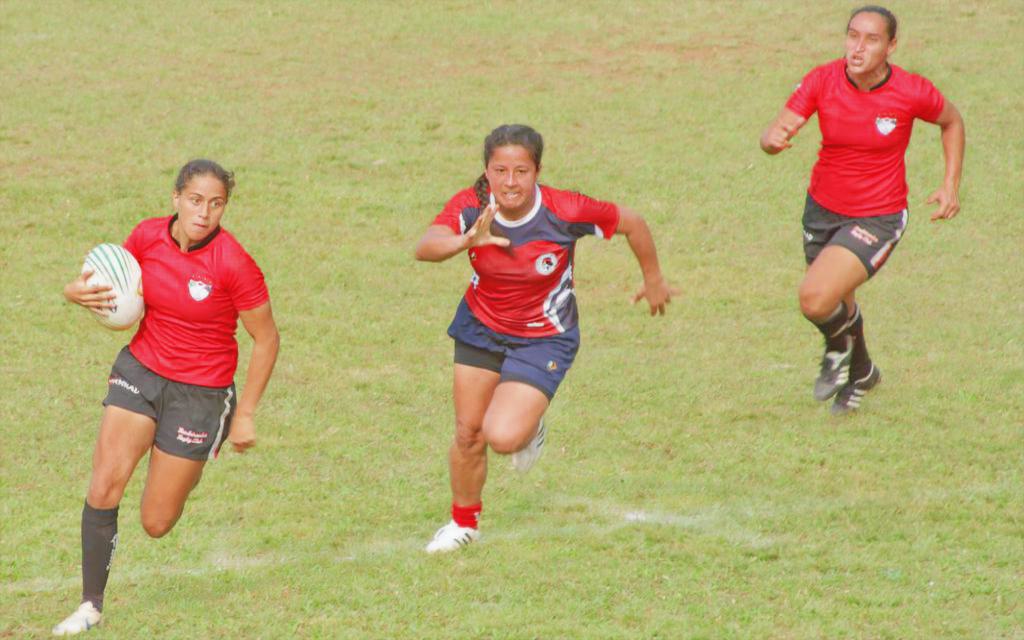}&
\includegraphics[width=0.23\linewidth,height=0.17\linewidth]{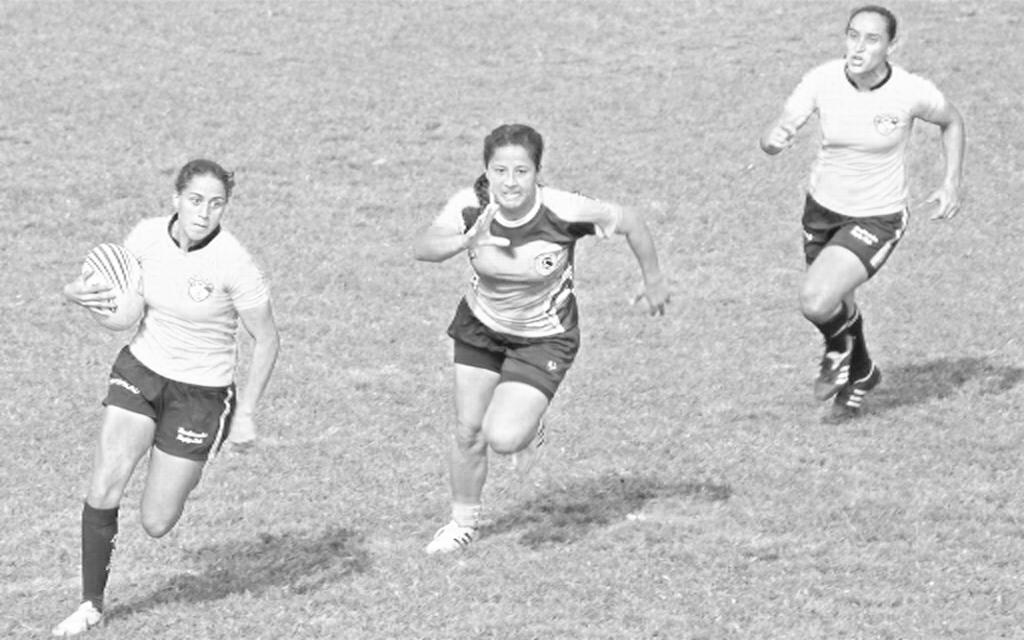}& 
\includegraphics[width=0.23\linewidth,height=0.17\linewidth]{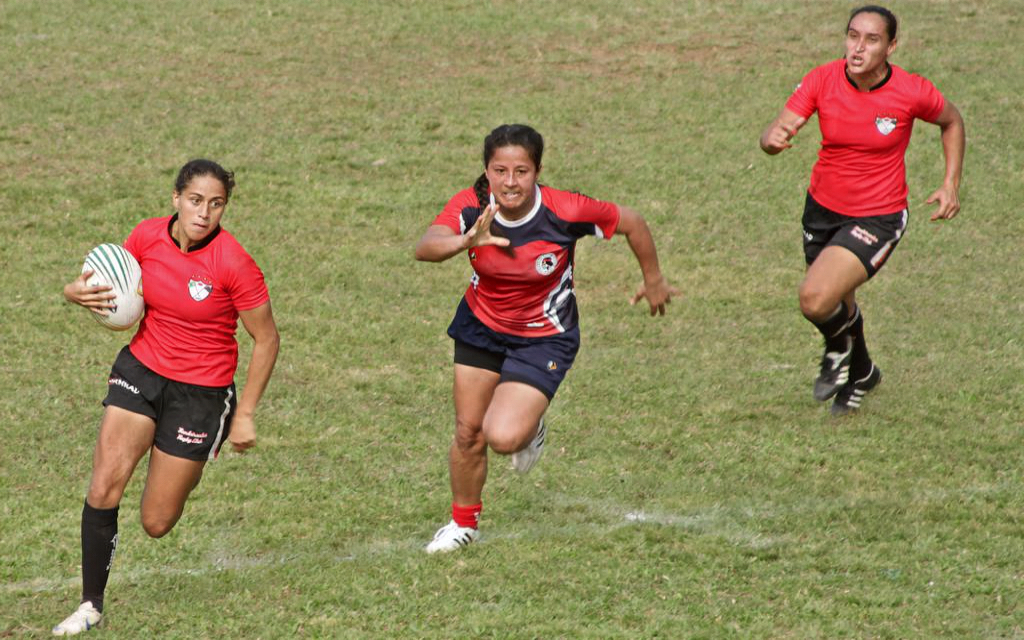}&

\\

\fontsize{8pt}{\baselineskip}\selectfont  (e)  SGNet&
\fontsize{8pt}{\baselineskip}\selectfont  (f) Ours-R &
\fontsize{8pt}{\baselineskip}\selectfont  (g) Ours-L &
\fontsize{8pt}{\baselineskip}\selectfont  (h) Ours \\

%
\end{tabular}
\end{center}
\caption{\ryn{Comparison of shadow removal results.} Existing methods (b-e) may \ryn{fail to completely remove the shadow} in the homogenous region and to recover the details in \ryn{the textured region}. Our method explicitly estimates the \yur{reflectance (f) and illumination \ryn{(g)} of the shadow image}, based on which we recast the lighting and correct \ryn{the texture in the shadow region}, resulting in a more visually pleasing prediction (h).}
\label{fig:teaser}
\end{figure}

Deep learning-based shadow removal methods can learn \ryn{the mapping} between shadow and 
\yu{shadow-free images} from \yu{large-scale} training data.
\yu{They \ryn{are typically based on}} different network structures and learning strategies (\eg, directional convolution~\cite{hu2019direction}, coarse-to-fine strategy~\cite{ding2019argan,wan2022style}, GANs~\cite{wang2018stacked,ding2019argan,cun2020towards}, and multi-exposure fusion~\cite{fu2021auto})
to learn color mapping directly, 
which may produce color-shifted artifacts~\cite{zhu2022bijective}. 
\yu{\ryn{Recent} methods \cite{chen2021canet,jin2021dc,zhu2022bijective} propose to model shadow-invariant color priors by, \eg, averaging the color of the whole image~\cite{chen2021canet}, using hand-crafted statistics~\cite{jin2021dc}, and training an auxiliary network~\cite{zhu2022bijective}.} 
Nonetheless, as shown in~\figref{fig:teaser}(b-e), although these \ryn{existing} methods may be able to recover the \ke{ lighting of shadow regions \rep{to some extent},} 
they fail to remove shadow remnants in the homogeneous region and to recover the details in the \ryn{texture region}.

\ryn{In this work, we observe that in a shadow region, the degradation
degree of object textures depends on the local illumination,
\ke{and enhancing only} the local illumination alone would not be able to fully recover the attenuated textures. There are two reasons for this.}
First, shadows may have sharp boundaries that {are mixed with object textures \ryn{and are difficult to be completely recovered simply by enhancing the lighting alone}.} Second, \ryn{object} textures may appear differently under different illuminations~\cite{serrano2021effect}.  
\ryn{Hence, unlike existing methods that attempt to directly recover the contents of the shadow regions,}
\ryn{in this paper, we propose to address this problem in two steps. First, we learn a color-to-illumination mapping, which helps regenerate the lighting in shadow regions. Second, we use the regenerated lighting to} guide the texture recovery.

\ryn{Based on this idea, we propose a novel shadow removal \ryn{method}, which \ryn{has two parts:}}
(1) a \textit{shadow-aware decomposition network} that explicitly estimates the illumination and reflectance layers for shadow images; \ryn{and} (2) a \textit{novel bilateral correction network} that first generates the homogeneous lighting and then recovers the textures \ryn{in shadow regions conditioned on the generated lighting}.
We follow the retinex theory~\cite{land1977retinex} to optimize the shadow-aware decomposition network
to ensure a physically-correct illumination estimation (\kk{see \figref{fig:teaser}(f,g), where \rynn{the} shadows are \rynn{learned to be captured by in the illumination layer only.}})
Our bilateral correction network has two novel designs: a local lighting correction \ryn{(LLC)} module and  an \yu{illumination-guided texture restoration \ryn{(IGTR)} module}. 
The former iteratively corrects the \rep{local} lighting of shadow regions by \rynnq{local \rep{conditional} denoising, 
 \yu{while the latter restores \yu{local textures} by %
scale-adaptive feature \yu{consistency} enhancement.}}
{As shown in~\figref{fig:teaser}(h), our method can remove the \ryn{shadow} and produce \ryn{a more accurate image}.}
\yu{In addition, as the \ke{widely used Shadow Removal Dataset (SRD)}~\cite{qu2017deshadownet} does not provide shadow masks, existing removal methods have to use shadow masks of different detection methods. For \ryn{fair evaluations}, we manually annotate the shadow masks for \ryn{it}.}

To sum up, we have the following key contributions:
\begin{itemize}
    \item \ryn{To remove shadows, we} propose to correct \yu{degraded} textures in shadow regions conditioned on \ryn{recovered} illumination. Our method includes a shadow-aware decomposition network and a novel bilateral correction network.

    \item We introduce two novel modules for the bilateral correction network: (1) a local lighting correction module that recasts shadow region lighting via \rep{local conditional denoising}, and (2) an illumination-guided texture restoration module that employs scale-adaptive features to enhance local textures, conditioned on \ryn{recovered} lighting. 
    \item We manually annotate accurate shadow masks for the SRD dataset, to ensure fair evaluation with existing methods, \rep{and  propel the advancement of this field. } 
    \item Extensive experiments on three shadow removal benchmarks 
    demonstrate that 
    \rep{(1) our method achieves state-of-the-art \ryn{performances}, and (2) our shadow-aware decomposition method can reduce the input requirement from a pixel-wise mask to a coarse bounding box.} 
\end{itemize}

\begin{figure*}[h!]
\centering
\includegraphics[width=0.85\linewidth]{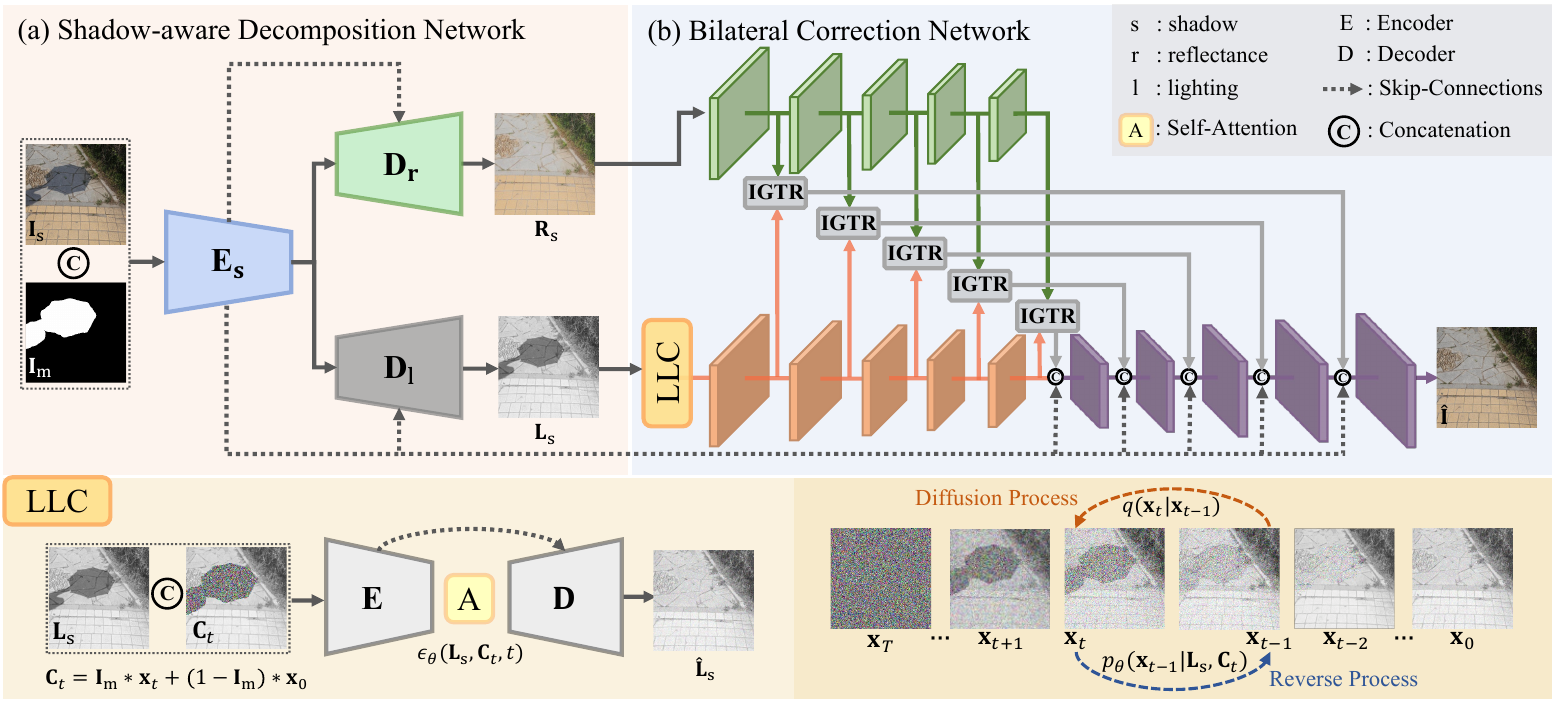}
   \caption{Method Overview. Given a shadow image $\mathbf{I}_{\text{s}}$ and a shadow mask $\mathbf{I}_{\text{m}}$ as input, the proposed method first  decomposes the shadow image into a reflectance layer $\mathbf{R}_{\text{s}}$ and an illumination layer $\mathbf{L}_{\text{s}}$ via the shadow-aware decomposition network. 
   $\mathbf{R}_{\text{s}}$, $\mathbf{L}_{\text{s}}$, and \yu{image features \ryn{through skip-connections}}
   are then fed into the bilateral correction network for lighting correction via the Local Lighting Correction (LLC) module \rep{to generate the shadow-free lighting $\hat{\mathbf{L}}_{\text{s}}$}, and texture restoration via the Illumination-Guided Texture Restoration (IGTR) \ryn{module}, \yu{and output the prediction $\mathbf{\hat{I}}$}. \rep{In LLC, \textit{t} is the time step, $\mathbf{x}_{0}$ is $\mathbf{L}_{\text{s}}$ \rynn{during} inference.}} 
\label{fig:pipeline}
\end{figure*}

\section{Related Work}
\label{sec:relatedwork}
{\bf Deep Shadow Removal Methods} take advantages of large-scale datasets~\cite{qu2017deshadownet,wang2018stacked,le2021physics}. 
Some methods focus on designing different network structures, \eg, contexts~\cite{qu2017deshadownet,cun2020towards}, directions~\cite{hu2019direction}, exposures~\cite{fu2021auto,Sun_2023_ICCV}, residuals~\cite{zhang2020ris} and structures~\cite{liu2023structure}, for shadow removal. Another group of works \cite{hu2019maskshadowgan,le2020from, liu2021from, jin2021dc, liu2021shadow, he2021unsupervised} propose a variety of learning strategies to train shadow removal networks using unpaired images. 
\rep{
\ryn{Recently}, several methods are proposed to model shadow-variant, \eg, SP+M+I~\cite{le2021physics} and EMNet~\cite{zhu2022efficient}, and shadow-invariant, \eg, CANet~\cite{chen2021canet}, DCGAN~\cite{jin2021dc} and BMNet~\cite{zhu2022bijective}, information to guide the shadow removal process.
SP+M+I proposes to estimate a group of linear parameters to represent the illumination information for shadow removal. 
EMNet further introduces non-linearity into the shadow formation model to predict a pixel-wise shadow degradation map. 
CANet removes shadows using a shadow-invariant color map obtained by averaging shadow image colors and transferring features between shadow and non-shadow regions.
DCGAN uses the shadow-invariant chromaticity map from traditional  methods~\cite{drew2003recovery,finlayson2009entropy} as pseudo labels for training.
Instead, BMNet directly trains a network to predict a shadow-invariant color map with the supervision of color maps averaged from shadow-free images and uses it to guide the removal process.
}
Unlike existing methods that \ke{implicitly} process lighting and textures simultaneously, we introduce a two-step approach to estimate and rectify the illumination in shadow regions first, followed by the restoration of degraded textures in these regions, conditioned on the recovered illumination.

\yu{
\textbf{Retinex Models}~\cite{land1977retinex}, which decompose an image into a reflectance image and an illumination image, provide a theoretical foundation for image formation and decomposition and have been widely used for image intrinsic decomposition~\cite{baslamisli2018cnn} and low-light  enhancement~\cite{wei2018deep,zhang2021beyond} tasks.
They typically design different network structures to estimate both reflection and illumination images, while Wang~\et~\cite{wang2019underexposed} assumes that the reflection image is the normal-light image and focuses on estimating only the enhanced illumination image.} 
However, these methods often fail to model illumination changes between shadow and non-shadow regions as they assume spatially consistent lighting. 
{In our work, we leverage various \yu{spatially-variant physical}  regularizations for modeling the lack of lighting in shadow regions during image decomposition.}

\textbf{Diffusion Models} are
generative models~\cite{sohl2015deep} that learn data distributions by the Gaussian noise blurring process and the reverse denoising process. They have been applied to \yu{various tasks}, \eg, image super-resolution~\cite{saharia2022image}, image generation~\cite{ho2020denoising,dhariwal2021diffusion} and color harmonization~\cite{Xu_2023_ICCV}. Some works~\cite{rombach2022high,saharia2022palette,zhang2023adding} propose to use 
additional inputs, \yu{\eg, texts, depth and sketch,} as conditions to enable global image generation or editing.
{In this work, we introduce the diffusion model into shadow removal and exploit its \rep{strong generative ability }  to recast the local lighting of shadow regions conditioned on the global lighting of shadow image.} 

\section{Proposed Method}
\label{sec:method}
Recent deep shadow removal methods typically formulate the task as $\mathbf{\hat{I}} = \phi(\mathbf{I}_{\text{s}} | \mathbf{P})$,
where $\phi(\cdot)$ is a pixel-to-pixel color mapping between a shadow image $\mathbf{I}_{\text{s}}$ and a shadow-free image 
\ke{$\mathbf{\hat{I}}$}. $\mathbf{P}$ represents the shadow position hints $\mathbf{I}_{\text{m}}$~(\eg, quadmap~\cite{wu2007natural} and  mask~\cite{le2019shadow}), which may additionally include the shadow invariant color map $\mathbf{I}_{\text{c}}$ (\eg, in \cite{chen2021canet} and \cite{jin2021dc}).
\ke{However, such a} formulation is not able to \ryn{recover} the degradation of lighting and textures in shadow regions separately. Although often less notable, shadows \ryn{often have clear} boundaries that may disrupt the original textures, and recovering the textures requires \ryn{correcting the local lighting first}.
\ke{Hence, instead of directly applying the above formula for shadow removal,}
in this paper, we propose to remove shadows by recasting the local lighting in shadow regions and correcting the textures conditionally.
Our method, depicted in \figref{fig:pipeline}, comprises two primary stages. In the first stage, we present a shadow-aware decomposition network for accurate reflectance-illumination separation. The second stage introduces a bilateral correction network that initially corrects degraded lighting in shadow areas using a local lighting correction module, followed by a progressive recovery of degraded texture details via an illumination-guided texture restoration module, conditioned on the corrected lighting.

\subsection{Shadow-aware Decomposition Network}
\label{subsec:stage1}

\textbf{Network Architecture.} 
As shown in \figref{fig:pipeline} (a), the \ke{proposed} shadow-aware decomposition network 
\ke{consists} of a shared encoder ($\mathbf{E}_{\text{s}}$) to extract shadow image features and two functionally distinct decoders ($\mathbf{D}_{\text{r}}$ and $\mathbf{D}_{\text{l}}$)
\ke{to} handle domain-specific reflectance and illumination features. 
The encoder has five convolutional layers, each of which employs the kernel size, stride, and padding of 4 and 2, and 1, respectively, and is followed by an InstanceNorm 
and a  Leaky-ReLU layer. The decoder  contains five transposed convolution layers with the same hyper-parameter settings as the convolutional layer in the encoder, each followed by an InstanceNorm 
and a ReLU.  
By default, skip connections are applied to all convolutional layers, where encoder and decoder features are concatenated. \
The decomposed reflectance $\mathbf{R}_{\text{s}}$ and illumination $\mathbf{L}_{\text{s}}$  are then normalized to a range of  $[0,1]$.%

\textbf{Shadow-aware Decomposition.} 
\ke{Optimizing $\mathbf{R}_{\text{s}}$ and $\mathbf{L}_{\text{s}}$ simultaneously is not straightforward, since}
there are no ground-truth reflectance and illumination in shadow removal. \ke{To this end, we design a new self-supervised learning strategy}. \ke{Specifically,} we leverage another network \footnote{We attach an all-zero map to the shadow-free image to keep the same input dimension as in the decomposition process of the shadow image. \rep{Note that this network is only used for training.}} \ke{of the same architecture to the shadow-aware decomposition network} to predict the reflectance $\mathbf{R}_{\text{sf}}$ and illumination $\mathbf{L}_{\text{sf}}$ of \ryn{the shadow-free image}. We train the two networks jointly with three physically correct \ke{self-supervised} regularizations to guide the shadow-aware decomposition process.

\textit{1) Maintaining Image Fidelity.}
We first \ke{apply} a $\mathit{L}_{1}$ loss to ensure that the decomposed layers can \ke{be reverted} to the original input (either shadow $\mathbf{I}_{\text{s}}$ or shadow-free $\mathbf{I}_{\text{sf}}$): 
\begin{equation}
\begin{aligned}
        \mathcal{L}_{\text {fid }}= \sum_{i \in \text {\{s,sf\} }} \left\|\mathbf{R}_i * \mathbf{L}_i-\mathbf{I}_i\right\|_1 \text {. }
\end{aligned}
\label{eq:recon_loss}
\end{equation}

\textit{2) Pulling Illumination Layers.} Although shadows may degrade both illumination and reflectance layers, the major difference between a shadow image and a shadow-free image should be preserved in their illumination layers.
\rep{Note that ground truth annotations for $\mathbf{R}$ and $\mathbf{L}$ are not available. Hence, we incorporate the Retinex theory~\cite{land1977retinex} into the illumination separation process by assuming consistent reflectance of shadow/non-shadow images:}
\begin{equation}
\begin{aligned}
        \mathcal{L}_{\text {ill }}=\left\|\mathbf{R}_{\text {s }}-\mathbf{R}_{\text {sf }}\right\|_1 
        + \sum_{i,j\in\text {\{s,sf\} }\text{,} i\neq j} \left\|\mathbf{R}_i * \mathbf{L}_j-\mathbf{I}_j\right\|_1 \text {, } 
\end{aligned}
\label{eq:R_loss}
\end{equation}
where the first term minimizes the differences between the reflectance layers of shadow and shadow-free images (\ie, $\mathbf{R}_{\text{s}}$ and $\mathbf{R}_{\text{sf}}$), and the second term  \rep{implicitly} minimizes the illumination difference of non-shadow regions between the shadow and shadow-free images (\ie, $\mathbf{L}_{\text{s}}$ and $\mathbf{L}_{\text{sf}}$). %

\textit{3) Constraints on Reflectance Layers.} 
Last, we apply the gradient constraints~\cite{meka2021real} on the reflectance layers to ensure \rep{texture preservation and color correction: }
\begin{equation}
\begin{aligned}
        \mathcal{L}_{\text {ref }} =\sum_{i\in\text { \{s,sf\} }} \left\|\nabla \mathbf{L}_\text{sf} * \exp \left(\lambda_{n} \nabla \mathbf{R}_i\right)\right\|_{1}\text{,} \\ \mathit{s.t.}\quad 0 \leq   \mathbf{R}_{i} < \mathbf{L}_{\text{sf}} \leq 1
\end{aligned}
\label{eq:recon_loss}
\end{equation}
where $\lambda_{\text{n}}$ is a hyper-parameter to adjust the weight {of the gradients of reflectance layers} and is set to $-20$. 
Note that we do not involve $\mathbf{L}_{\text{s}}$ in \kk{\ryn{Eq.~\ref{eq:recon_loss}}}, to avoid the illumination layer degrading into a shadow matte~\cite{qu2017deshadownet} and the reflectance layer being identical to the input shadow image.

\begin{figure}[t!]
\begin{center}
\begin{tabular}{c@{\hspace{0.3mm}}c@{\hspace{0.3mm}}c@{\hspace{0.3mm}}c@{\hspace{0.3mm}}c@{\hspace{0.3mm}}c}
\includegraphics[width=0.187\linewidth,height=0.13\linewidth]{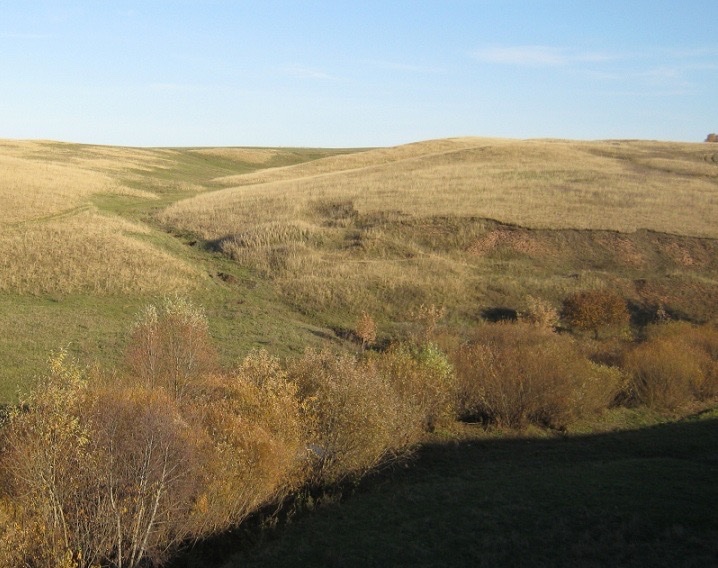}&
\includegraphics[width=0.187\linewidth,height=0.13\linewidth]{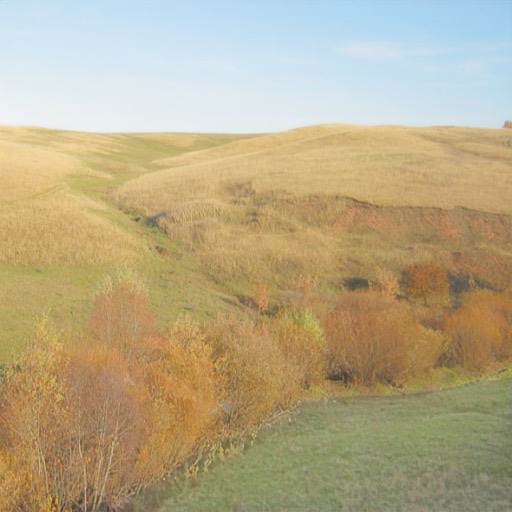}&
\includegraphics[width=0.187\linewidth,height=0.13\linewidth]{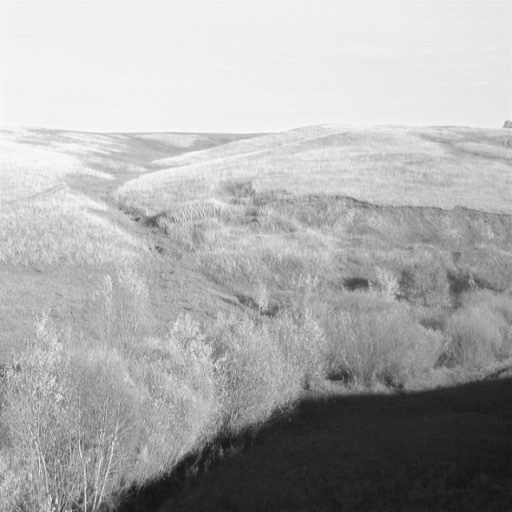}& 
\includegraphics[width=0.187\linewidth,height=0.13\linewidth]{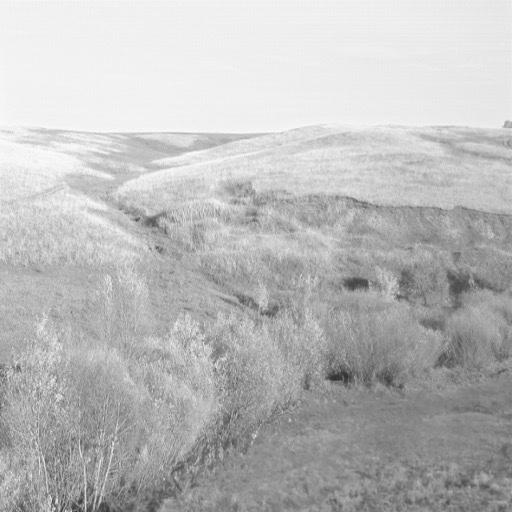}&
\includegraphics[width=0.187\linewidth,height=0.13\linewidth]{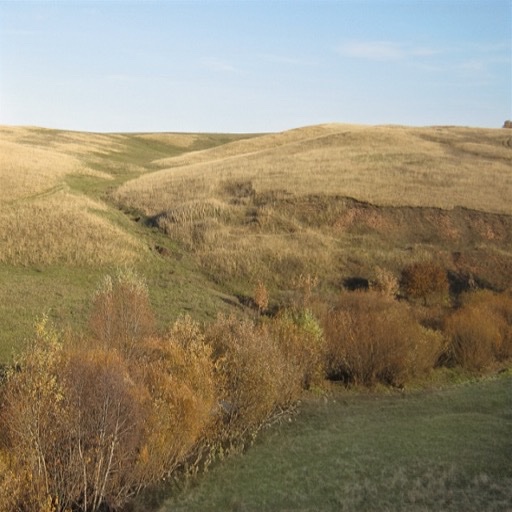}& \\
\includegraphics[width=0.187\linewidth,height=0.13\linewidth]{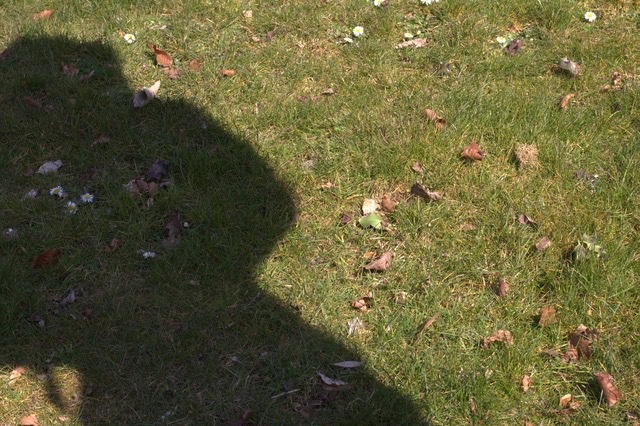}&
\includegraphics[width=0.187\linewidth,height=0.13\linewidth]{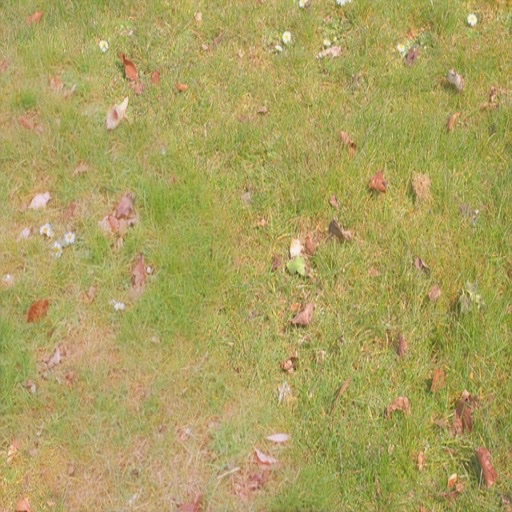}&
\includegraphics[width=0.187\linewidth,height=0.13\linewidth]{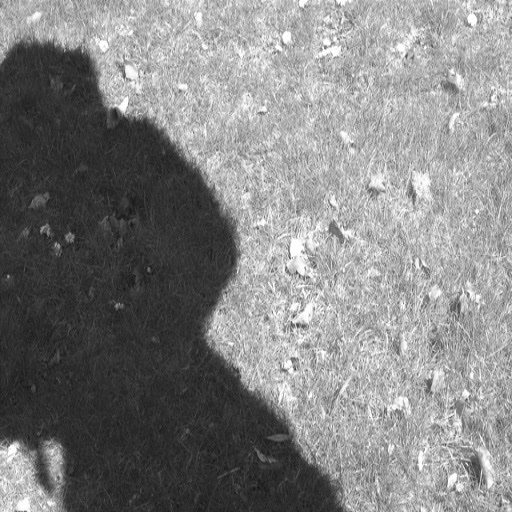}& 
\includegraphics[width=0.187\linewidth,height=0.13\linewidth]{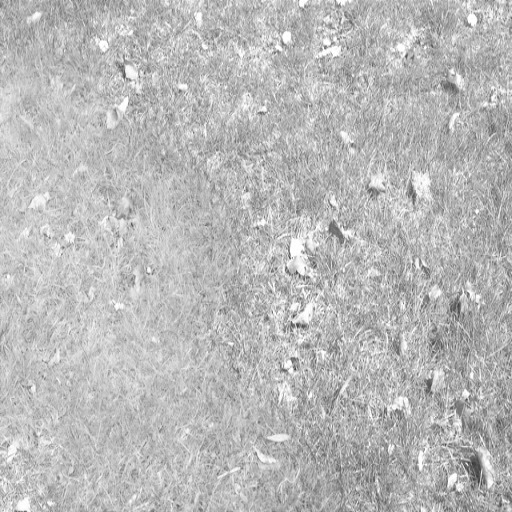}&
\includegraphics[width=0.187\linewidth,height=0.13\linewidth]{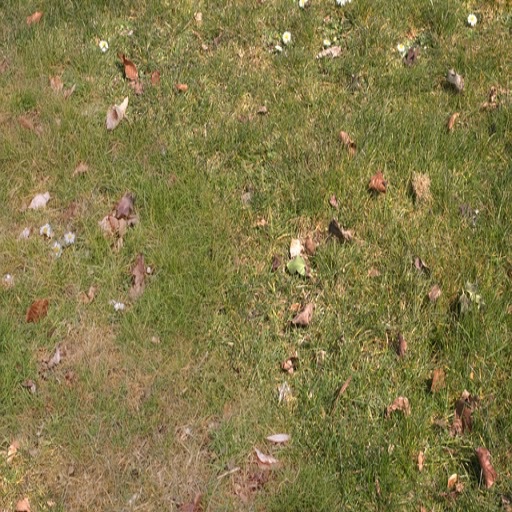}&  \\

\fontsize{8pt}{\baselineskip}\selectfont (a) $\mathbf{I}_{\text{s}}$ &
\fontsize{8pt}{\baselineskip}\selectfont (b) $\mathbf{R}_{\text{s}}$ &
\fontsize{8pt}{\baselineskip}\selectfont (c) $\mathbf{L}_{\text{s}}$ &
\fontsize{8pt}{\baselineskip}\selectfont (d) $\hat{\mathbf{L}}_{\text{s}}$ &
\fontsize{8pt}{\baselineskip}\selectfont (e) $\hat{\mathbf{I}}_{\text{s}}$ \\
\end{tabular}
\end{center}
\caption{\rep{Two examples  of our shadow-aware decomposition and final prediction results in  real-world samples.}}
\label{fig:decomp_case}
\end{figure}

The whole shadow-aware decomposition process is supervised by the following loss function:
\begin{equation}
\begin{aligned}
        \mathcal{L}_{\text{de}}=  \mathcal{L}_{\text{fid}} +  \mathcal{L}_{\text{ill}}  + w_{\text{r}} \mathcal{L}_{\text{ref}}\text {, }
\end{aligned}
\label{eq:decomp_loss}
\end{equation}
where $w_{\text{r}}$ is balancing parameter and empirically set to 0.1. See \figref{fig:decomp_case}  for our shadow-aware decomposition illustration.

\subsection{Bilateral  Correction Network}
\label{subsec:ban}
With decomposed illumination \ke{$\mathbf{L}_{\text{s}}$} and reflectance \ke{$\mathbf{R}_{\text{s}}$}, we first recast regional lighting via the proposed {\it local lighting correction (LLC) module} to produce a homogeneous illumination layer.
We then extract cross-level features of the reflectance and re-casted illumination via an encoder,
and apply the proposed {\it \ryn{illumination-guided texture restoration (IGTR)} module} {at multiple scales} to enhance their feature consistency and progressively restore local textures. 

\textbf{Local Lighting Correction.} 
Since local illumination of an image is sample-specific and spatially non-uniform~\cite{zhu2022efficient}, learning a fixed set of parameters for all samples through a regression-based network is typically insufficient and inaccurate. 
To solve such a problem, we consider local lighting correction as a generation problem and resort to the diffusion model to recast the lighting iteratively.

\yu{However, we note that DDPM is essentially a global denoising process, which is not able to focus on the local shadow regions of our task.
Hence, we formulate our local lighting correction module based on the DDPM with two conditions:
the independent shadow lighting condition \rep{$\mathbf{L}_{s}$} and the \rep{time-}embedded non-shadow lighting condition \rep{$\mathbf{C}_{t}$}.
The former focuses our module on the local lighting of shadow regions, while the latter provides globally-consistent lighting guidance for the local light generation: }
\begin{equation}
    \begin{aligned}
        \mathbf{C}_t = \mathbf{I}_{\text{m}} * \mathbf{x}_{t} + \left(\mathbf{1} -\mathbf{I}_{\text{m}}\right) * \mathbf{x}_{0} \text{,}
    \end{aligned}
    \label{eq: x_t_hat}
\end{equation}
\rep{where $t$ is the time step, $\mathbf{x}_{t} = \left(\sqrt{\bar{\alpha}_t} \mathbf{x}_0+\sqrt{1-\bar{\alpha}_t} \boldsymbol{\epsilon}_{t}\right)$ in which  $\bar{\alpha}$ is the variance schedule, and $\boldsymbol{\epsilon}$ is \rynn{the} randomly sampled gaussian noise. $\mathbf{1}$ is a mask filled with 1. 
During training, we set $\mathbf{x}_{0}$ to $\mathbf{L}_{\text{sf}}$ and feed the conditions and $t$ to the noise prediction network $\boldsymbol{\epsilon}_\theta\left( \cdot \right)$ to conduct local conditional noise prediction and update its parameters.
During testing, we set $\mathbf{x}_{0}$ to $\mathbf{L}_{\text{s}}$ and perform the iterative 
local conditional denoising  to generate the $\hat{\mathbf{L}}_{\text{s}}$.
}
We adopt an improved UNet~\cite{dhariwal2021diffusion}
as our noise prediction network $\boldsymbol{\epsilon}_{\theta}(\cdot)$, 
and  train it \rep{using the MSE denoising loss} within the  shadow regions, as:
\begin{equation}
\begin{aligned}
        \mathcal{L}_{\mathrm{denoise}}=\mathbb{E}_{\boldsymbol{\epsilon} \sim \mathcal{N}(\mathbf{0}, \mathbf{I})}\left[\mathbf{I}_{\text{m}} * \left\|\boldsymbol{\epsilon}-\boldsymbol{\epsilon}_\theta\left(\mathbf{C}_{t} \text{,}\; \mathbf{L}_{\text{s}} \text{,}\;  t\right)\right\|_2^2\right] \text{.}
\end{aligned}
\label{eq:diff_loss}
\end{equation}
In this way, the diffusion model can  focus on the shadow regions conveniently and exploit the correct illumination of the non-shadow regions. 
\ryn{Refer to} the comparison between the 3rd and 4th columns in~\figref{fig:decomp_case} for an \yu{visual} illustration.
\begin{figure}
\centering
\includegraphics[width=0.9\linewidth]{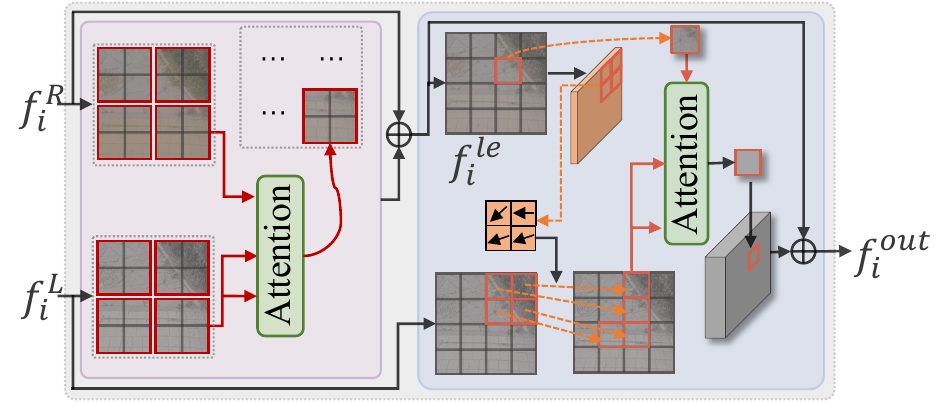}
   \caption{Overview of the proposed Illumination-Guided Texture Restoration (IGTR) module. It aims to correct the textures with the guidance of \ryn{the recovered} local lighting.
   }
\label{fig:IGTR}
\end{figure}

\textbf{Illumination-Guided Texture Restoration.} 
With the re-generated illumination $\hat{\mathbf{L}}_{\text{S}}$, we propose 
to condition the local texture restoration on the recovered local lighting, with two kinds of lighting-to-texture correspondences being modeled for texture fidelity.
We first correct each local region of the reflectance layer according to the lighting of the corresponding local region in the illumination layer. Then, we enhance the texture consistency by learning the correspondence between each local reflectance \yur{token} and adjacent regions in the illumination layer.

 \figref{fig:IGTR} shows the overview of the proposed IGTR module.
Formally, given the features of reflectance and corrected illumination at \yu{scale} $i\in\{1\text{,}2\text{,} \cdots \text{,}5\}$ as $f^{\text{R}}_{i}\in R^{H_{i} \times W_{i} \times C_{i}}$ and $f^{\text{L}}_{i}\in R^{H_{i} \times W_{i} \times C_{i}}$ (\ryn{where} $H$, $W$ and $C$ represent height, width, and channel), we first divide them into local regions of size $K_{i}\times K_{i}$.
\ryn{We then compute} the \yu{local lighting-to-texture correspondence} between each corresponding region in \yu{$f^{\text{R}}_{i}$ and $f^{\text{L}}_{i}$} via co-attention (CoA)~\cite{vaswani2017attention}:
\begin{equation}
\begin{aligned}
    \text{CoA}(f^{\text{R}}_{i}\text{, } f^{\text{L}}_{i}) = \mathcal{S}\left((\text{W}_q f^{\text{R}}_{i}) (\text{W}_k{f^{\text{L}}_{i}})^{T} / \sqrt{\text{d}}\right) (\text{W}_v f^{\text{L}}_{i})\text{,}
\end{aligned}
\label{eq:igtr_attn}
\end{equation}
where  $\text{d}$ and $\mathcal{S}$ are the scaling factor and SoftMax operation, and  $\text{W}_{j}\text{, } j\in\{\text{q}\text{, } \text{k}\text{, }\text{v}\}$ is  projection function (\rep{conv. layer}) to reduce the feature dimension  by half. 
We \yu{then obtain} the enhanced texture features $f^{\text{le}}_{i}  = \text{CoA}(f^{\text{R}}_{i}\text{, }f^{\text{L}}_{i}) + f^{\text{R}}_{i} $.

We further build the non-local lighting-to-texture correspondence between each \yur{token} in $f^{\text{R}}_{i}$ and the neighbouring regions in $f^{\text{L}}_{i}$,
where the adjacent regions in  $f^{\text{L}}_{i}$ are adaptively searched based on enhanced texture features $f^{\text{le}}_{i}$: 
\begin{equation}
\begin{aligned}
    \hat{f}^{\text{L}}_{i} = \mathcal{B}(f^{\text{L}}_{i} \text{,}\; \mathbf{Shift}(f^{\text{le}}_{i}))\text{,}
\end{aligned}
\label{eq:igtr_da_detail}
\end{equation}
where  $\mathbf{Shift}$ is a shifting network \cite{xia2022vision} to learn offsets for each location in the local region, \rep{and $\mathcal{B}(\cdot)$ is the bilinear interpolation for feature-resampling.}
We then obtain the final enhanced texture feature: $f^{\text{out}}_{i} = \text{CoA}(f^{\text{le}}_{i}\text{,} \hat{f}^{\text{L}}_{i}) + f^{\text{le}}_{i} $.

\textbf{Loss Functions.} 
We adopt the L1 loss and the perceptual loss~\cite{johnson2016perceptual} for training the texture restoration process, as:
\begin{equation}
\begin{aligned}
        \mathcal{L}_{\text {re }}=  \left\|\mathbf{\hat{I}}\text{, } \mathbf{I}_{\text{sf}}\right\|_1 + \lambda_{\text{vgg}} * \left\|\text{VGG}(\mathbf{\hat{I}}) \text{, } \text{VGG}(\mathbf{I}_{\text{sf}})\right\|_1 \text {. }
\end{aligned}
\label{eq:texture_loss}
\end{equation}
where $\lambda_{\text{vgg}}$ is empirically set to 0.1 \rep{to maintain the same gradient magnitude of \rynn{the} two loss items.}

\section{Ground Truth Labeling on SRD}

SRD~\cite{qu2017deshadownet}, the inaugural large-scale dataset for shadow removal, doesn't offer ground truth shadow masks. Hence, extant removal techniques employ various methods like shadow detector~\cite{zhu2022bijective}, Otsu's algorithm and morphology~\cite{fu2021auto}, or shadow matting with threshholding~\cite{cun2020towards} to generate shadow masks. 
\tableref{tab:mask_comp} assesses the accuracy of these shadow masks against our labeled masks as ground truth, using shadow detection metrics (PER and BER~\cite{zhu2021mitigating}). 
It uncovers considerable quality variations, often rendering evaluations of shadow removal methods on the SRD biased and potentially impacting removal performance~\cite{zhu2022bijective}.
Thus, to ensure a fair evaluation on this dataset, we manually annotate the pixel-wise shadow masks for the SRD dataset.

\begin{table}
\begin{center}
\footnotesize
		\renewcommand\tabcolsep{7.5pt}
		\renewcommand\arraystretch{0.5}
            \begin{tabular}{l|c|c|c|c|c}
            \toprule
         \multicolumn{2}{l|}{Metrics} & MTMT&  DHAN &FDR & AEF \\
         \midrule
         \multicolumn{2}{l|}{PER~$\downarrow$} &20.35 & 20.80 & 15.03 & 12.03 \\
         \multicolumn{2}{l|}{BER~$\downarrow$} & 11.81& 10.87 & 10.19 & 6.49  \\
         \bottomrule
            \end{tabular}
    \end{center}
    \caption{Existing shadow mask quality varies significantly when assessed against our labeled masks as ground truth, with lower PER and BER values indicating higher quality.}
\label{tab:mask_comp}
\end{table}

\section{Experiments}
\label{sec:experiments}

{\bf Implementation Details.} 
Our method is implemented via the PyTorch Toolbox on a single NVIDIA TESLA V100 GPU with 32G memory, is optimized using the Adam \cite{diederik2015adam} optimizer. The initial learning rate, $\beta_{1}$, $\beta_{2}$, and batch size being set to 0.0002, 0.9, 0.999, and 12.
Learning rate adjustment utilizes a warmup and cosine decay strategy. 
For the local diffusion process, we set the times steps $T$, initial and end variance scheduler~$\beta_{\text{t}}$ to $\{1000\text{,}0.0001\text{,}0.02\}$ and $\{50\text{,}0.0001\text{,}0.5\}$ for training and testing. 
Data is augmented by random flipping and cropping, and resized to 256$\times$256 for training. Shadow-aware decomposition and bilateral correction networks are trained for 100k and 200k iterations, respectively.

{\bf Datasets.} 
We conduct %
experiments on three shadow removal datasets, \ie, SRD~\cite{qu2017deshadownet}, ISTD \cite{wang2018stacked}, and ISTD+ \cite{le2021physics}. 
SRD consists of 3,088 paired shadow and shadow-free images, which are split into {2680} for training and {408} for testing.
ISTD contains 1,870 shadow images, shadow masks, and shadow-free image triplets, of which 1,330 are used for training and 540 for testing.
ISTD+ further corrects the color inconsistency problem of images from the ISTD.

{\bf Evaluation Metrics.} We follow \cite{le2021physics} to compute the root mean square error (RMSE) between the  results and ground truth shadow-free images in the LAB color space, 
and report the peak signal-to-noise ratio (PSNR) and structural similarity (SSIM) for comparisons. 
\rep{All metrics are \rynn{computed based on the} 256 resolution.}

\begin{table}[t!]
\begin{center}
\footnotesize
		\renewcommand\tabcolsep{1.2pt}
		\renewcommand\arraystretch{1.}
            \begin{tabular}{l|ccc|ccc|ccc}
            \toprule
         \multirow{2}{*}{Methods} &\multicolumn{3}{c|}{RMSE~$\downarrow$} &\multicolumn{3}{c|}{PSNR~$\uparrow$} &\multicolumn{3}{c}{SSIM~$\uparrow$} \\
          & \textit{S} & \textit{NS} & \textit{All} & \textit{S} & \textit{NS} & \textit{All} & \textit{S} & \textit{NS} & \textit{All} \\
         \midrule
         DSC &17.25&16.58&16.76&26.71&24.70&21.63&0.914&0.756&0.657 \\
         DHAN &7.53&3.55&4.61&33.81&35.02&30.74&0.979&0.982&0.958 \\
         AEF &8.13&5.57&6.25&33.26&30.39&27.96&0.970&0.938&0.902 \\
         DCGAN&8.03&3.82&4.94&33.36&34.87&30.56&0.973&0.980&0.947 \\
        EMNet &9.55&6.67&7.43&30.24&26.32&24.16&0.940&0.851&0.779 \\
         BMNet &7.11&3.11&4.18&34.82&36.54&31.97&0.981&\textbf{0.986}&0.965 \\
         SGNet &7.45&3.05&4.23&33.76&36.48&31.39&0.979&0.984&0.960\\
         Ours &\textbf{5.49}&\textbf{3.00}&\textbf{3.66}&\textbf{36.51}&\textbf{37.71}&\textbf{33.48}&\textbf{0.983}&\textbf{0.986}&\textbf{0.967}\\
         
         \bottomrule
            \end{tabular}
    \end{center}
\caption{Quantitative comparisons with state-of-the-art shadow removal methods on the SRD dataset. All methods are \yu{tested} using our manually annotated shadow masks. The best results are marked in  bold. S, NS, and ALL indicate the shadow regions, non-shadow regions, and the whole image. }
\label{tab:sota_srd}
\end{table}

\begin{table}[t!]
\begin{center}
\footnotesize
		\renewcommand\tabcolsep{1.3pt}
		\renewcommand\arraystretch{1.}
            \begin{tabular}{l|ccc|ccc|ccc}
            \toprule
         \multirow{2}{*}{Methods} &\multicolumn{3}{c|}{RMSE~$\downarrow$} &\multicolumn{3}{c|}{PSNR~$\uparrow$} &\multicolumn{3}{c}{SSIM~$\uparrow$} \\
          & \textit{S} & \textit{NS} & \textit{All} & \textit{S} & \textit{NS} & \textit{All} & \textit{S} & \textit{NS} & \textit{All} \\
         \midrule
         ST&9.99&6.05&6.65&33.74&29.51&27.44&0.981&0.958&0.929 \\
         DSC &8.72&5.04&5.59&34.64&31.26&29.00&0.984&0.969&0.944 \\
         DHAN &8.26&5.56&6.37&34.65&29.81&28.15&0.983&0.937&0.913 \\
         DCGAN&11.43&5.81&6.57&31.69&28.99&26.38&0.976&0.958&0.922 \\
         G2R &10.72&7.55&7.85&31.63&26.19&24.72&0.975&0.967&0.932\\
         AEF &7.91&5.51&5.88&34.71&28.61&27.19&0.975&0.880&0.945 \\
         EMNet &7.78 & 4.72 & 5.22 & 36.27 & 31.85 & 29.98 & 0.986 & 0.965 & 0.944 \\
         BMNet &7.60&4.59&5.02&35.61&32.80&30.28&\textbf{0.988}&0.976&0.959 \\
         Ours &\textbf{6.54}&\textbf{3.40}&\textbf{3.91}&\textbf{36.61}&\textbf{35.75}&\textbf{32.42}&\textbf{0.988}&\textbf{0.979}&\textbf{0.961}\\
         
         \bottomrule
            \end{tabular}
    \end{center}
\caption{Quantitative comparisons with state-of-the-art shadow removal methods on the ISTD dataset.}

\label{tab:sota_istd}
\end{table}

\begin{table}[h!]
\begin{center}
\footnotesize
		\renewcommand\tabcolsep{1.3pt}
		\renewcommand\arraystretch{1.0}
            \begin{tabular}{l|ccc|ccc|ccc}
            \toprule
         \multirow{2}{*}{Methods} &\multicolumn{3}{c|}{RMSE~$\downarrow$} &\multicolumn{3}{c|}{PSNR~$\uparrow$} &\multicolumn{3}{c}{SSIM~$\uparrow$} \\
          & \textit{S} & \textit{NS} & \textit{All} & \textit{S} & \textit{NS} & \textit{All} & \textit{S} & \textit{NS} & \textit{All} \\
         \midrule
          P+M+D & 9.67&2.82&3.94&33.09&35.35&30.15&0.983&0.978&0.951\\

         DCGAN& 10.41&3.63&4.74&32.00&33.56&28.77&0.976&0.968&0.932\\
         G2R &7.35&2.91&3.64&35.78&35.64&31.93&0.987&0.977&0.957\\
         AEF &6.55&3.77&4.22&36.04&31.16&29.45&0.978&0.892&0.861 \\
         SP+M+I&5.91&2.99&3.46&37.60&36.02&32.94&\textbf{0.990}&0.976&0.962 \\
         BMNet&6.10 &2.90&3.50&\rep{37.30}&\rep{37.93} &\rep{33.95}&\textbf{0.990}&\rep{0.981}&\rep{0.965} \\
         SGNet & 5.93& 2.92&3.41&36.79&35.57&32.45&\textbf{0.990}&0.977&0.962 \\
         Ours& \textbf{5.69}&\textbf{2.31}&\textbf{2.87}&\textbf{38.04}&\textbf{39.15}&\textbf{34.96}&\textbf{0.990}&\textbf{0.984}&\textbf{0.968}\\
         \bottomrule
            \end{tabular}
    \end{center}
\caption{Quantitative  comparison with state-of-the-art shadow removal methods on the ISTD+ dataset.} 
\label{tab:sota_ISTD+}
\end{table}

\begin{figure*}[htp]
    \begin{center}
    \scalebox{0.88}{
    \begin{tabular}
    {c@{\hspace{0.85mm}}c@{\hspace{0.85mm}}c@{\hspace{0.85mm}}c@{\hspace{0.85mm}}c@{\hspace{0.85mm}}c@{\hspace{0.85mm}}c@{\hspace{0.85mm}}c@{\hspace{0.85mm}}c@{\hspace{0.85mm}}c}

    \includegraphics[width=0.12\linewidth,height=0.07\linewidth]{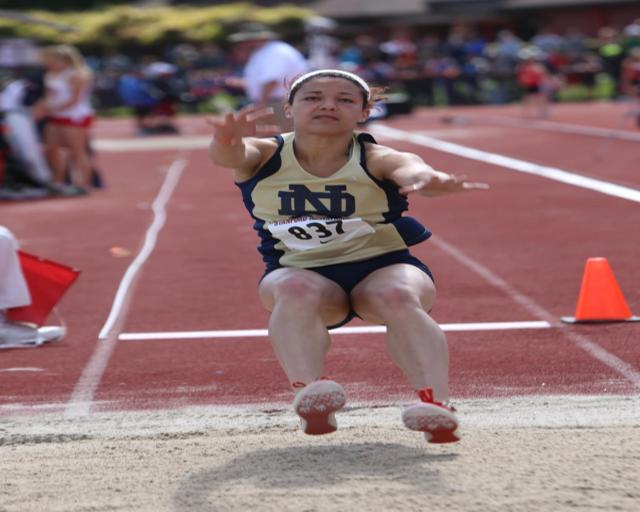}&
    \includegraphics[width=0.12\linewidth,height=0.07\linewidth]{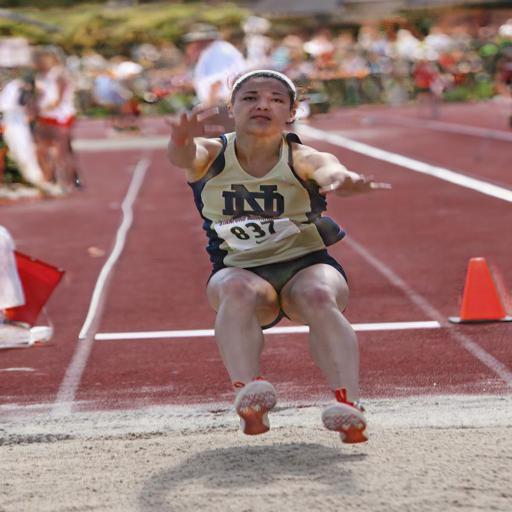}& 
    \includegraphics[width=0.12\linewidth,height=0.07\linewidth]{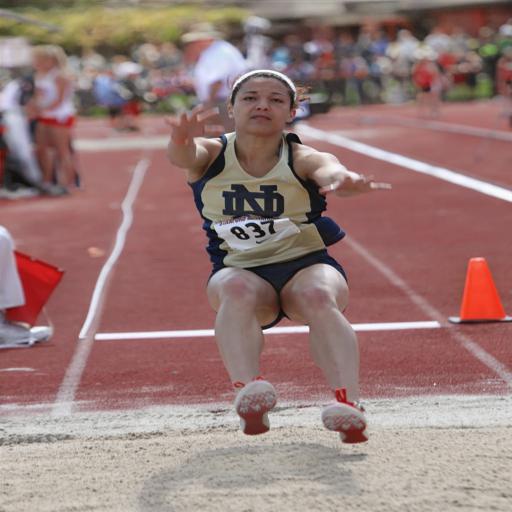}& 
    \includegraphics[width=0.12\linewidth,height=0.07\linewidth]{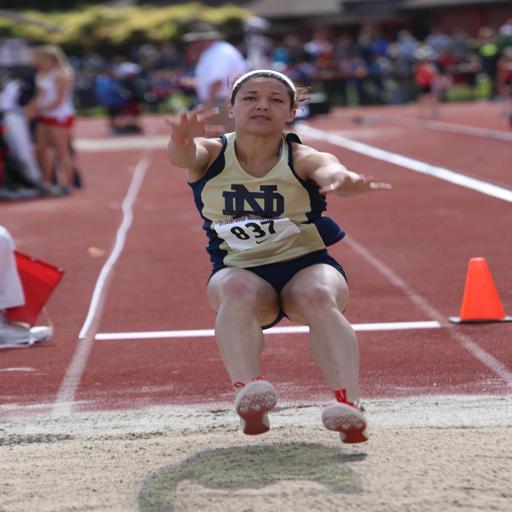}& 
    \includegraphics[width=0.12\linewidth,height=0.07\linewidth]{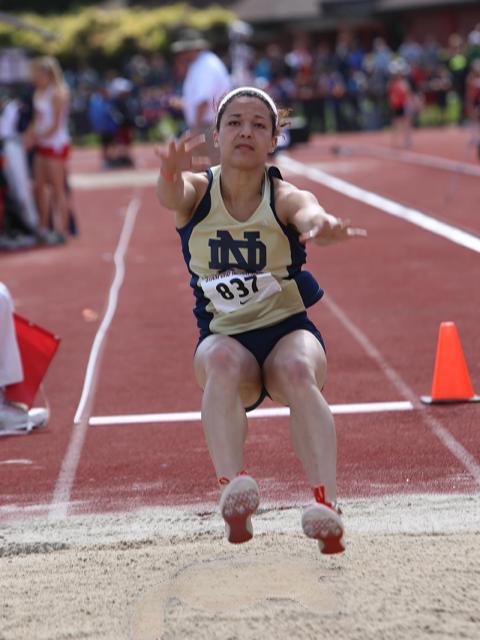}&
    \includegraphics[width=0.12\linewidth,height=0.07\linewidth]{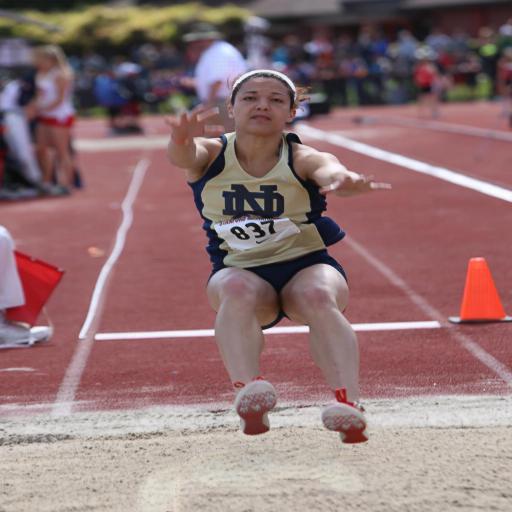}&
    \includegraphics[width=0.12\linewidth,height=0.07\linewidth]{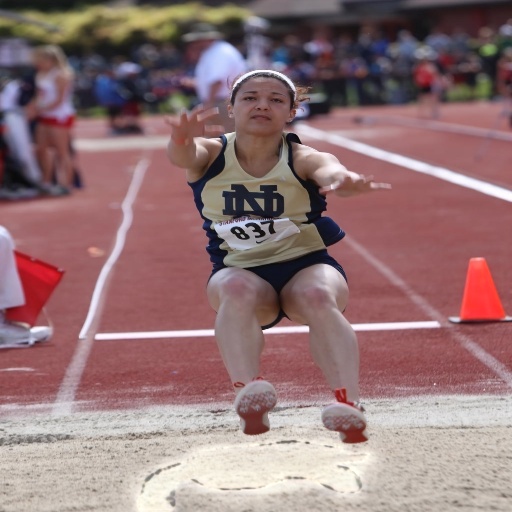}& 
    \includegraphics[width=0.12\linewidth,height=0.07\linewidth]{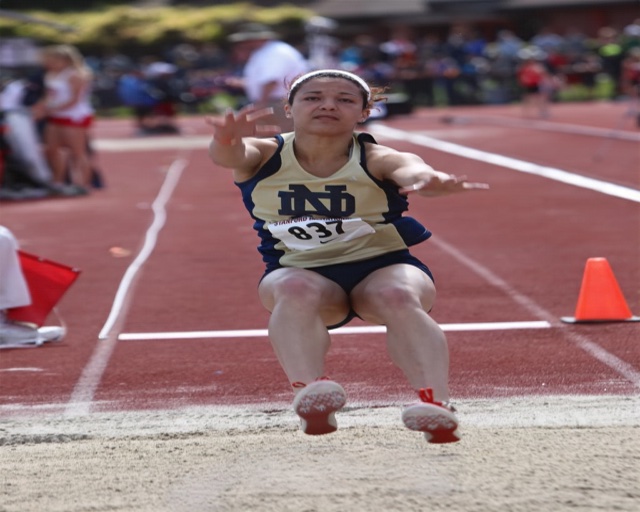}& \\
        
    \includegraphics[width=0.12\linewidth,height=0.07\linewidth]{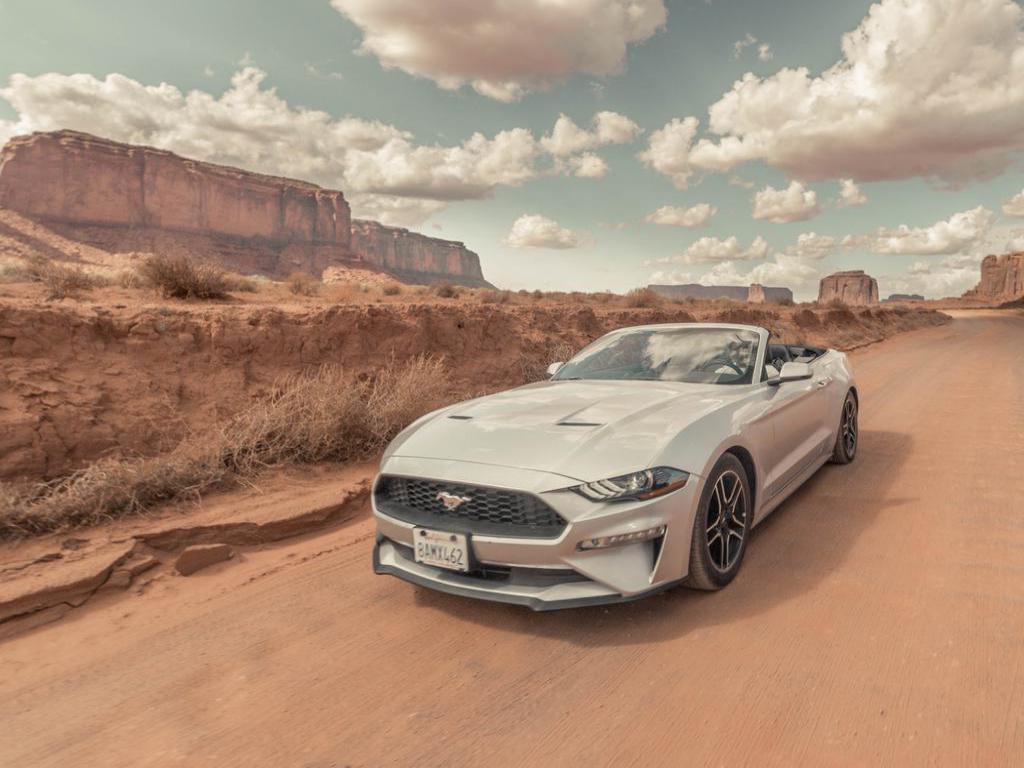}&
    \includegraphics[width=0.12\linewidth,height=0.07\linewidth]{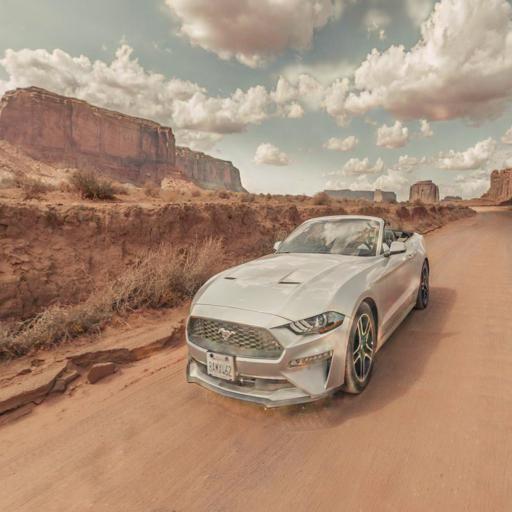}& 
    \includegraphics[width=0.12\linewidth,height=0.07\linewidth]{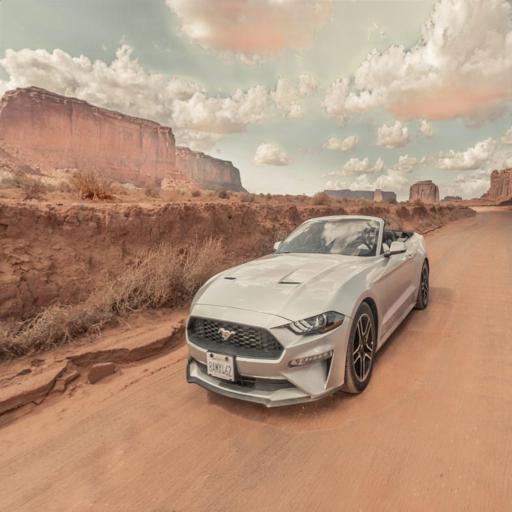}& 
    \includegraphics[width=0.12\linewidth,height=0.07\linewidth]{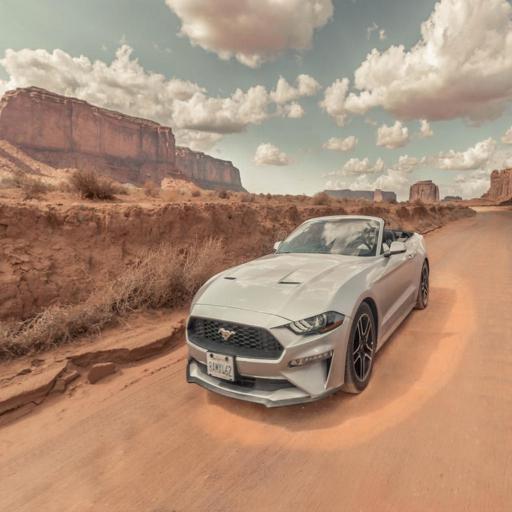}& 
    \includegraphics[width=0.12\linewidth,height=0.07\linewidth]{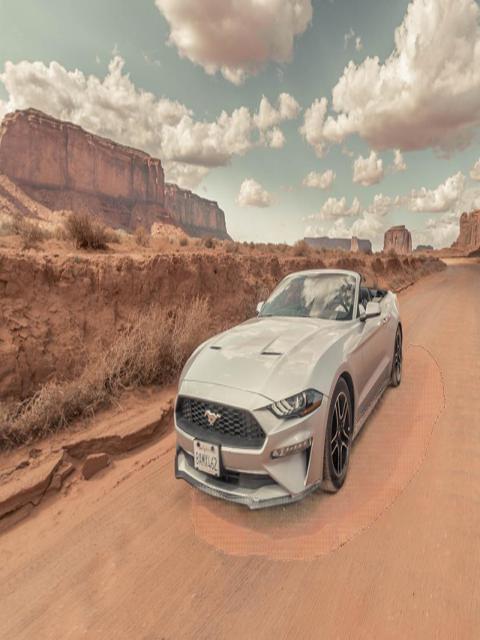}&
    \includegraphics[width=0.12\linewidth,height=0.07\linewidth]{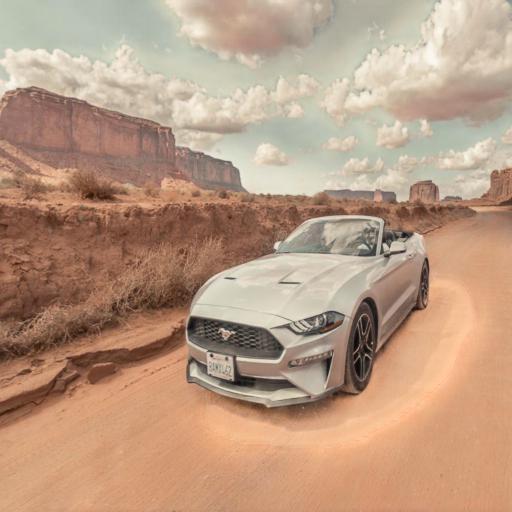}&
    \includegraphics[width=0.12\linewidth,height=0.07\linewidth]{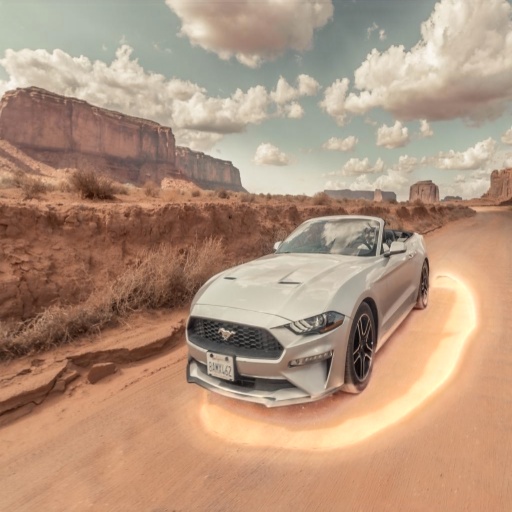}&
    \includegraphics[width=0.12\linewidth,height=0.07\linewidth]{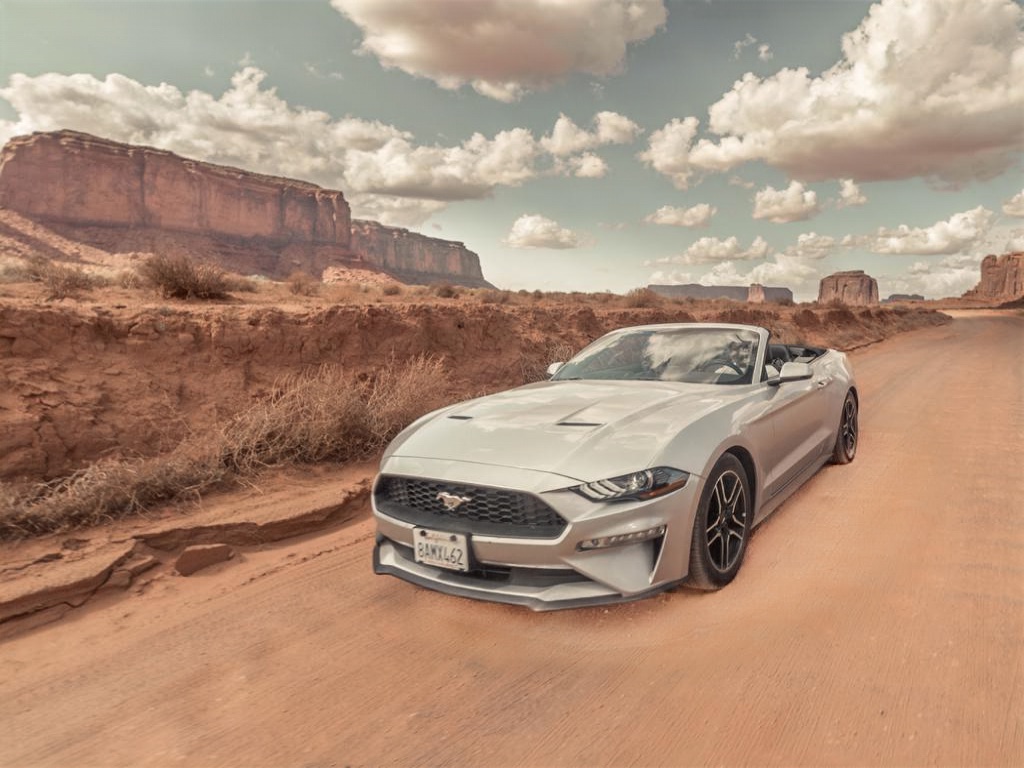}& \\

    \includegraphics[width=0.12\linewidth,height=0.07\linewidth]{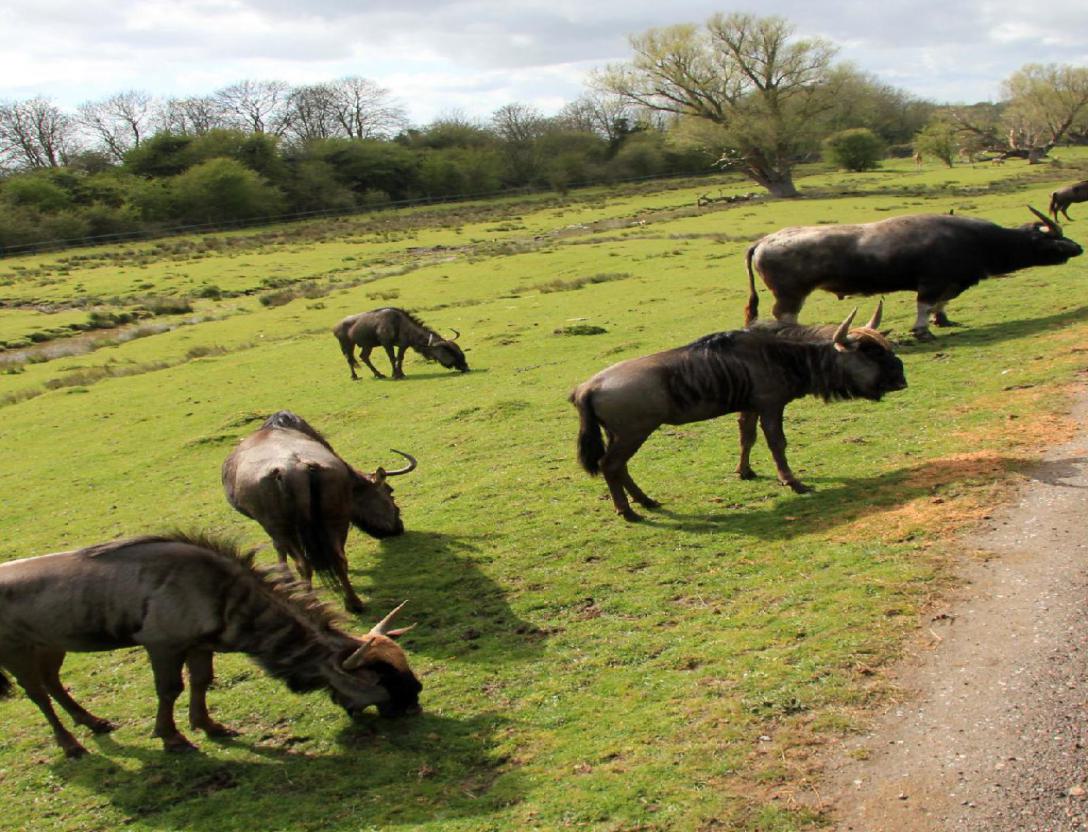}&
    \includegraphics[width=0.12\linewidth,height=0.07\linewidth]{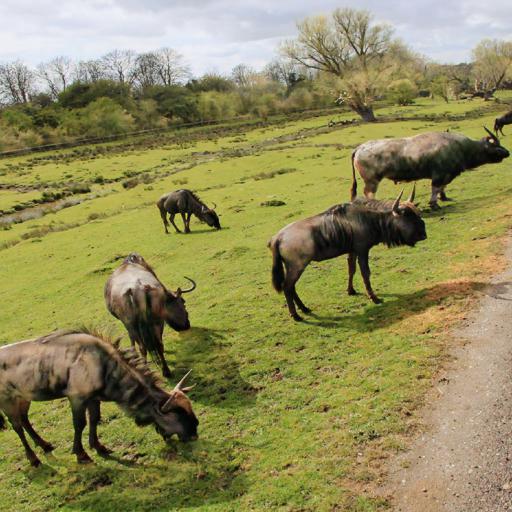}& 
    \includegraphics[width=0.12\linewidth,height=0.07\linewidth]{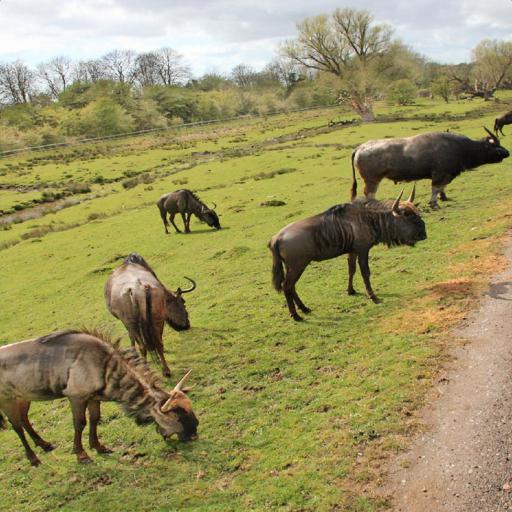}& 
    \includegraphics[width=0.12\linewidth,height=0.07\linewidth]{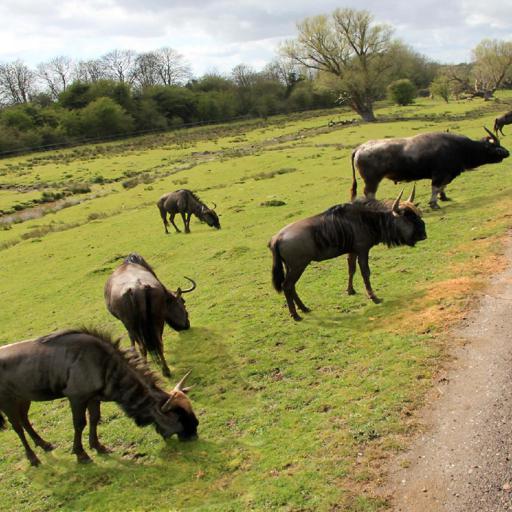}& 
    \includegraphics[width=0.12\linewidth,height=0.07\linewidth]{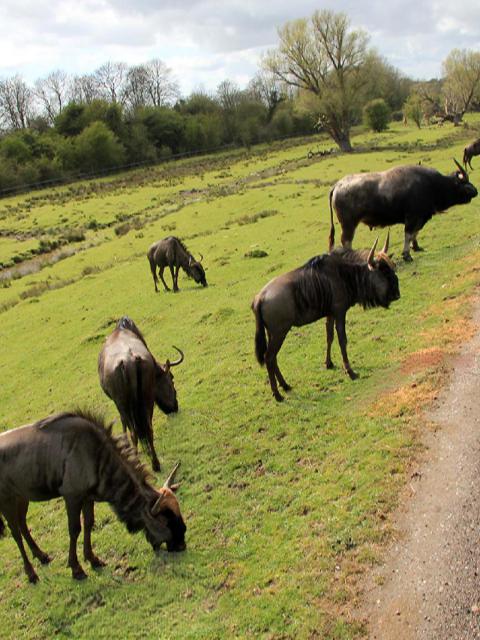}&
    \includegraphics[width=0.12\linewidth,height=0.07\linewidth]{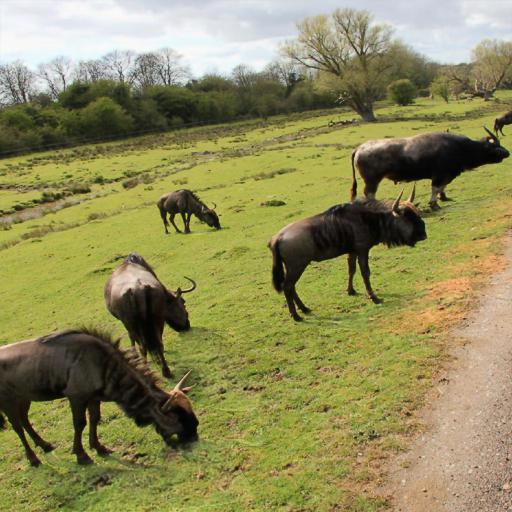}&
    \includegraphics[width=0.12\linewidth,height=0.07\linewidth]{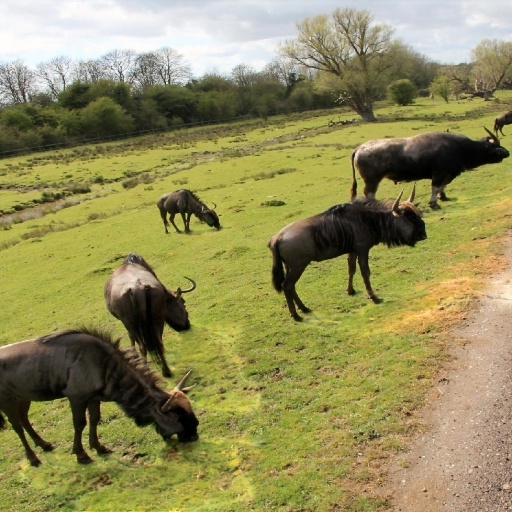}& 
    \includegraphics[width=0.12\linewidth,height=0.07\linewidth]{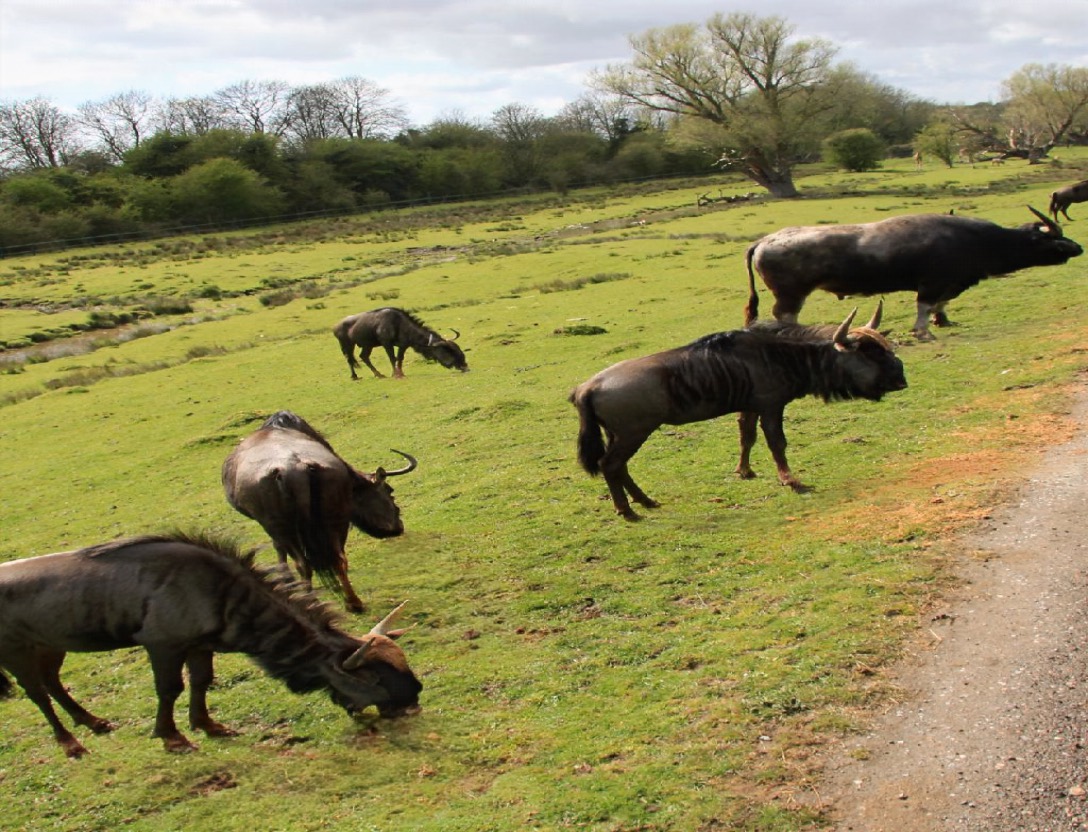}&  \\

    \includegraphics[width=0.12\linewidth,height=0.07\linewidth]{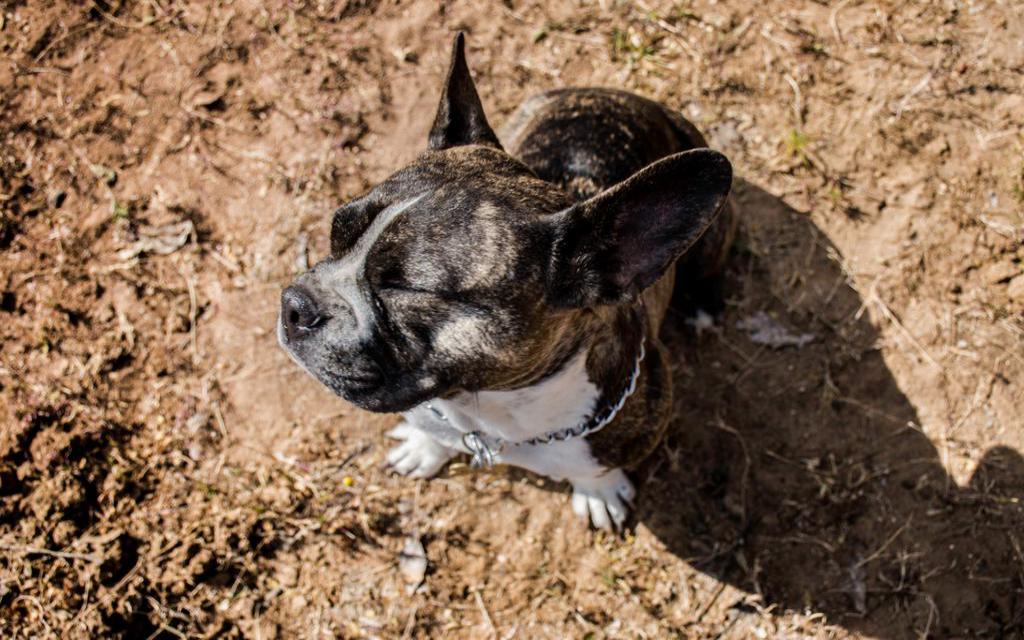}&
    \includegraphics[width=0.12\linewidth,height=0.07\linewidth]{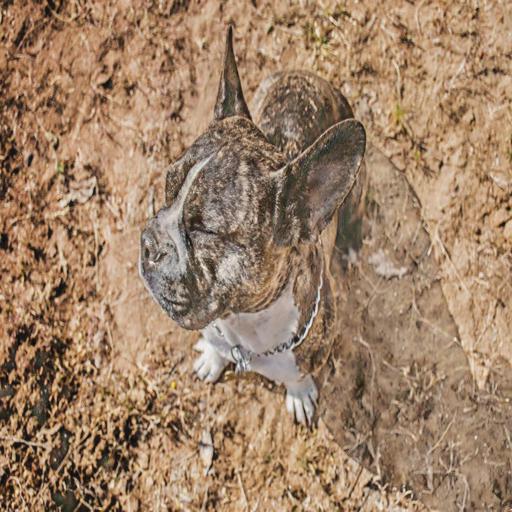}& 
    \includegraphics[width=0.12\linewidth,height=0.07\linewidth]{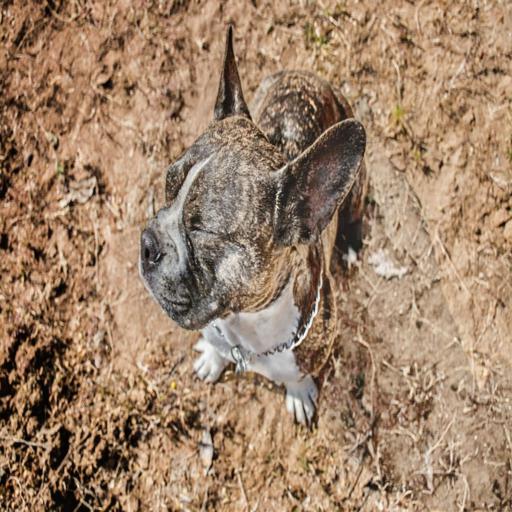}& 
    \includegraphics[width=0.12\linewidth,height=0.07\linewidth]{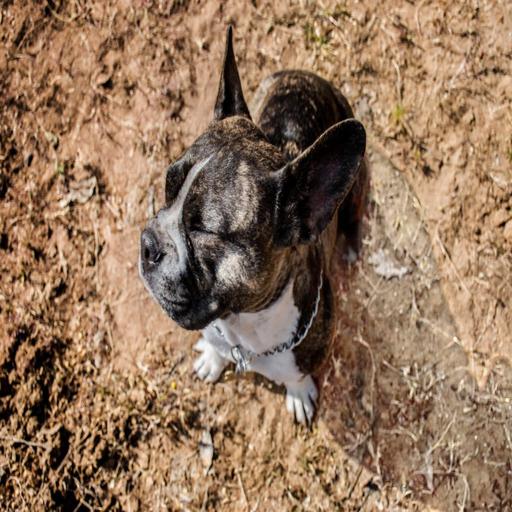}& 
    \includegraphics[width=0.12\linewidth,height=0.07\linewidth]{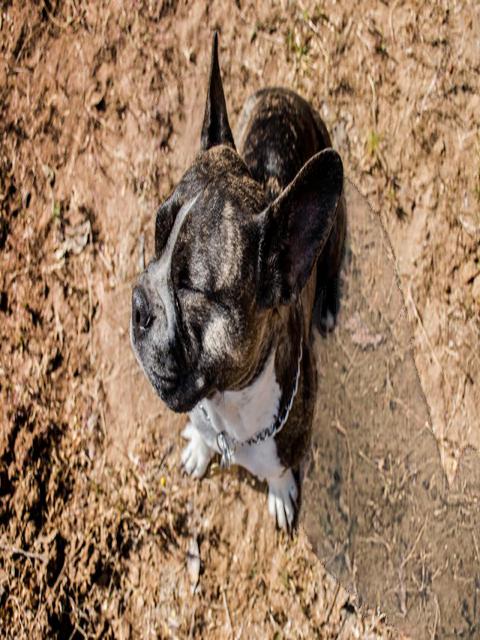}&
    \includegraphics[width=0.12\linewidth,height=0.07\linewidth]{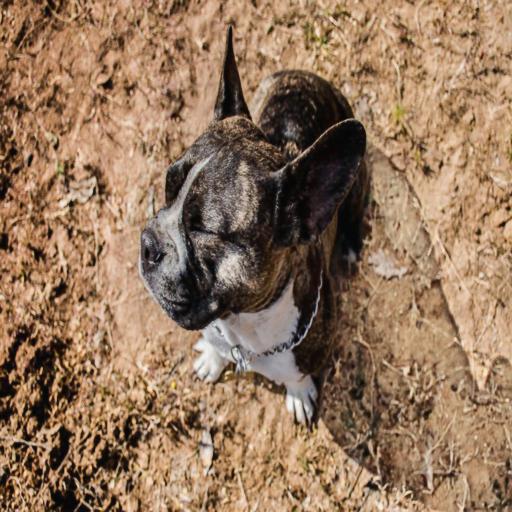}&
    \includegraphics[width=0.12\linewidth,height=0.07\linewidth]{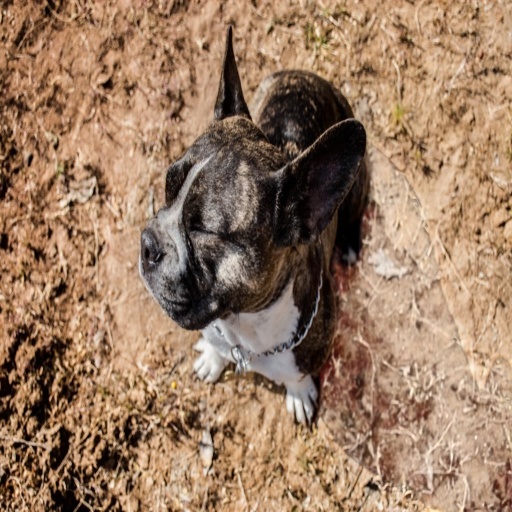}& 
    \includegraphics[width=0.12\linewidth,height=0.07\linewidth]{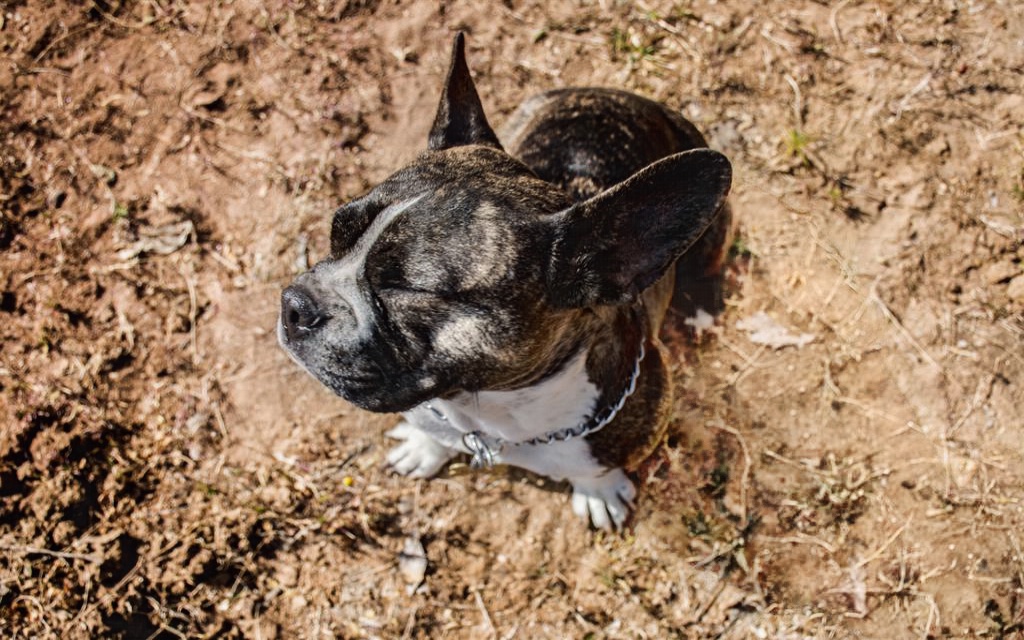}& \\

    \fontsize{9pt}{\baselineskip}\selectfont Input Shadow &
    \fontsize{9pt}{\baselineskip}\selectfont DC-GAN  & 
    \fontsize{9pt}{\baselineskip}\selectfont  DHAN  &
    \fontsize{9pt}{\baselineskip}\selectfont AEF   & 
    \fontsize{9pt}{\baselineskip}\selectfont G2R  &
    \fontsize{9pt}{\baselineskip}\selectfont SGNet   &
    \fontsize{9pt}{\baselineskip}\selectfont  BMNet  &
    \fontsize{9pt}{\baselineskip}\selectfont  Ours  & \\
    \end{tabular}
    }
    \end{center}
    \caption{Visual comparisons with state-of-the-art shadow removal methods on real-world samples. }
    \label{fig:sota_comp}
    \end{figure*}

\subsection{Comparing to State-of-the-arts}
\label{sub:comp_sota}

We compare our method with twelve shadow removal methods: 
ST-CGAN~\cite{wang2018stacked},
DSC~\cite{hu2019direction},
DHAN~\cite{cun2020towards},
P+M+D~\cite{le2020from}, 
DCGAN~\cite{jin2021dc}, SP+M+I~\cite{le2021physics},
G2R~\cite{liu2021from},
AEF~\cite{fu2021auto}, 
EMNet~\cite{zhu2022efficient}, 
BMNet~\cite{zhu2022bijective} and SGNet~\cite{wan2022style}. 

For SRD (\tableref{tab:sota_srd}), our method considerably surpasses other methods, even beating the latest color mapping-based method SGNet~\cite{wan2022style} and shadow-invariant map-based method BMNet~\cite{zhu2022bijective} by 26.3$\%$  and 22.8$\%$ in RMSE for shadow regions, respectively.
When assessed on ISTD and ISTD+ (\tableref{tab:sota_istd} and 4), our method again outperforms, achieving a 13.9$\%$ and 6.7$\%$ decrease in RMSE for shadow regions compared to BMNet~\cite{zhu2022bijective}. 
As shown in \figref{fig:sota_comp}, visual comparisons on real-world samples also illustrate our method's proficiency in reducing shadow ghosting (row-1), maintaining color consistency (row-2), and correcting textures (row-3 and 4) after shadow removal, particularly in challenging scenarios with homogenous colors or textured scenes. 

\subsection{Internal Analysis}
\label{sec:ablation}

\begin{figure}[t!]
\begin{center}
\begin{tabular}{c@{\hspace{0.3mm}}c@{\hspace{0.3mm}}c@{\hspace{0.3mm}}c@{\hspace{0.3mm}}c@{\hspace{0.3mm}}c}
\includegraphics[width=0.187\linewidth,height=0.13\linewidth]{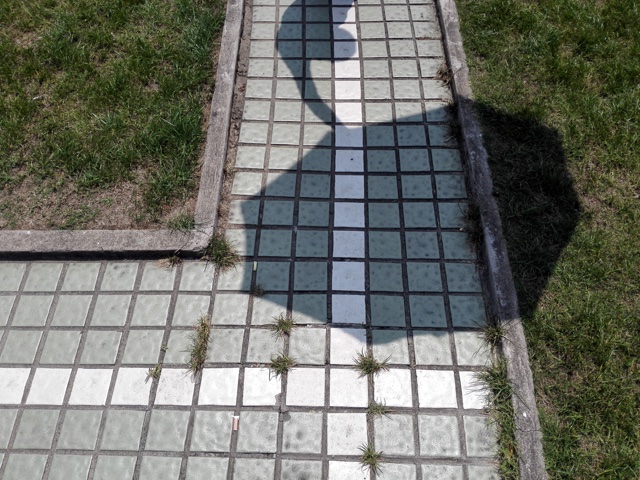}&
\includegraphics[width=0.187\linewidth,height=0.13\linewidth]{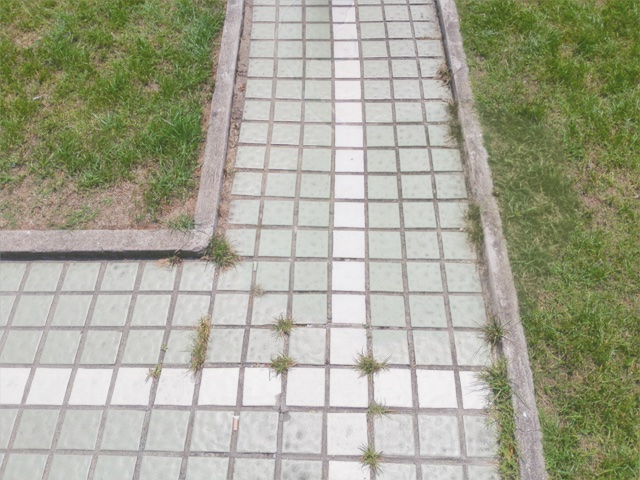}&
\includegraphics[width=0.187\linewidth,height=0.13\linewidth]{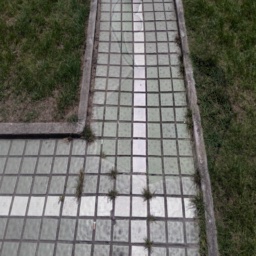}&
\includegraphics[width=0.187\linewidth,height=0.13\linewidth]{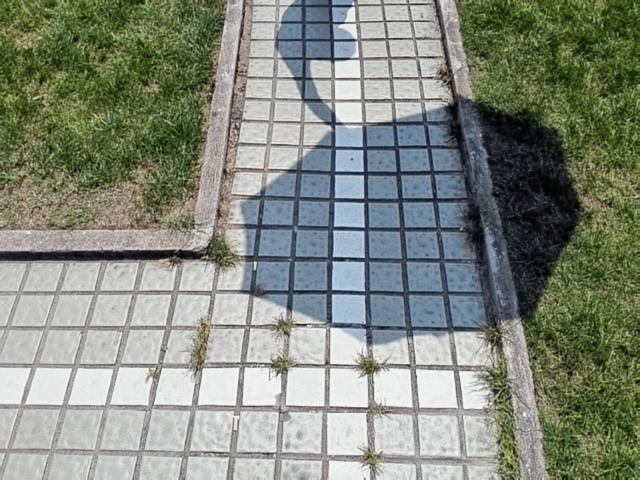}& 
\includegraphics[width=0.187\linewidth,height=0.13\linewidth]{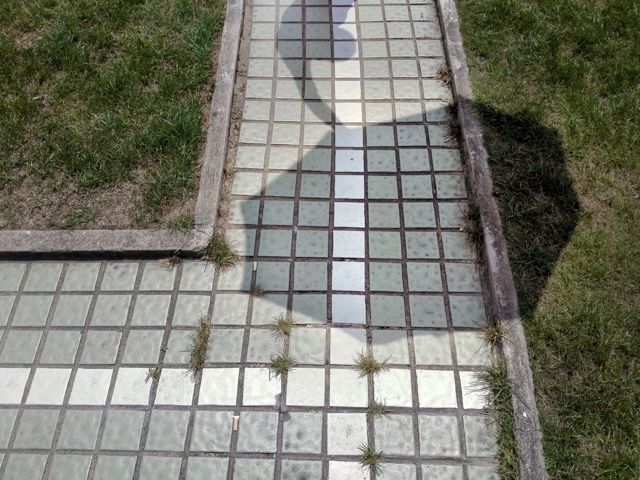}& 
\\
\includegraphics[width=0.187\linewidth,height=0.13\linewidth]{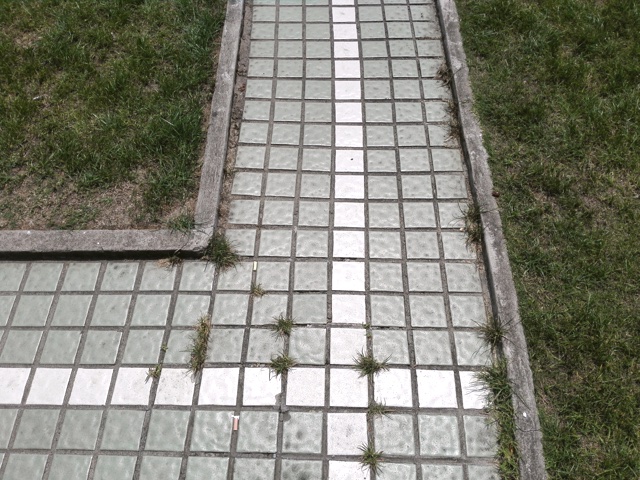}&
\includegraphics[width=0.187\linewidth,height=0.13\linewidth]{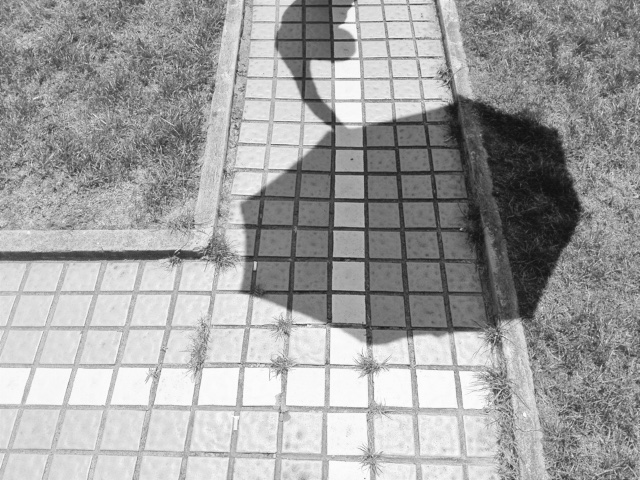}&
\includegraphics[width=0.187\linewidth,height=0.13\linewidth]{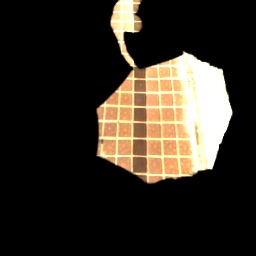}&
\includegraphics[width=0.187\linewidth,height=0.13\linewidth]{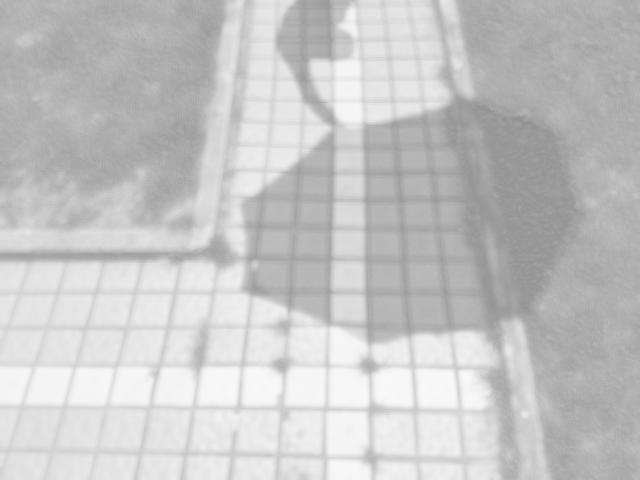}& 
\includegraphics[width=0.187\linewidth,height=0.13\linewidth]{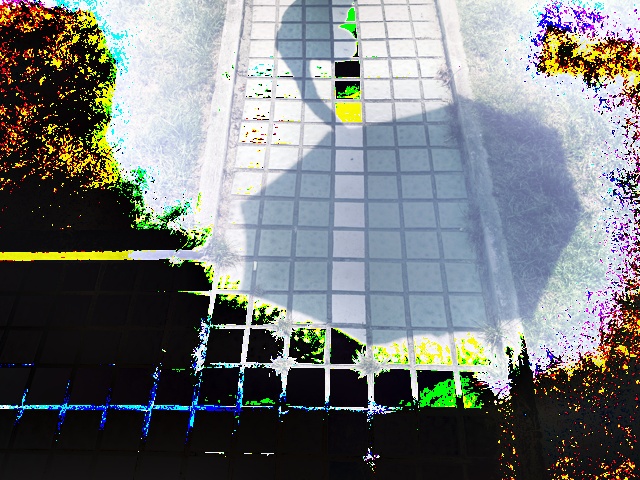}& 
\\
\fontsize{8pt}{\baselineskip}\selectfont In \& GT &
\fontsize{8pt}{\baselineskip}\selectfont  Ours& 
\fontsize{8pt}{\baselineskip}\selectfont  EMNet &  
\fontsize{8pt}{\baselineskip}\selectfont  RetinexNet& %
\fontsize{8pt}{\baselineskip}\selectfont  DeepUPE& %
\\
\end{tabular}
\end{center}
\caption{Visual comparisons of decomposition results among our method, shadow removal method EMNet, and \rep{two retinex-based low-light enhancement methods RetinexNet~\cite{wei2018deep} and DeepUPE~\cite{wang2019underexposed} re-trained using shadow \rynn{masks} as additional input.}}%
\label{fig:decomp_comp}
\end{figure}

\textbf{Shadow-aware Decomposition Evaluation.}
We illustrate the correctness of our shadow-aware decomposition  by visually comparing it to related  methods, as there is no ground truth for quantitative evaluation.
\figref{fig:decomp_comp} compares ours to the shadow formation-based EMNet and two retinex-based methods, DeepUPE~\cite{wang2019underexposed} and RetinexNet~\cite{wei2018deep}.
The degradation map of EMNet jointly models the lighting and reflectance, resulting in the pinkish remnants in their shadow-free image. 
On the other hand, the two retinex-based methods fail to separate the reflectance and illumination layers \rep{due to their spatially invariant property}. 
In contrast, our spatially-variant regularizations ensure the physically correct  shadow decomposition. %

\begin{figure}[t!]
\begin{center}
\begin{tabular}{c@{\hspace{0.3mm}}c@{\hspace{0.3mm}}c@{\hspace{0.3mm}}c@{\hspace{0.3mm}}c@{\hspace{0.3mm}}c}
\includegraphics[width=0.187\linewidth,height=0.13\linewidth]{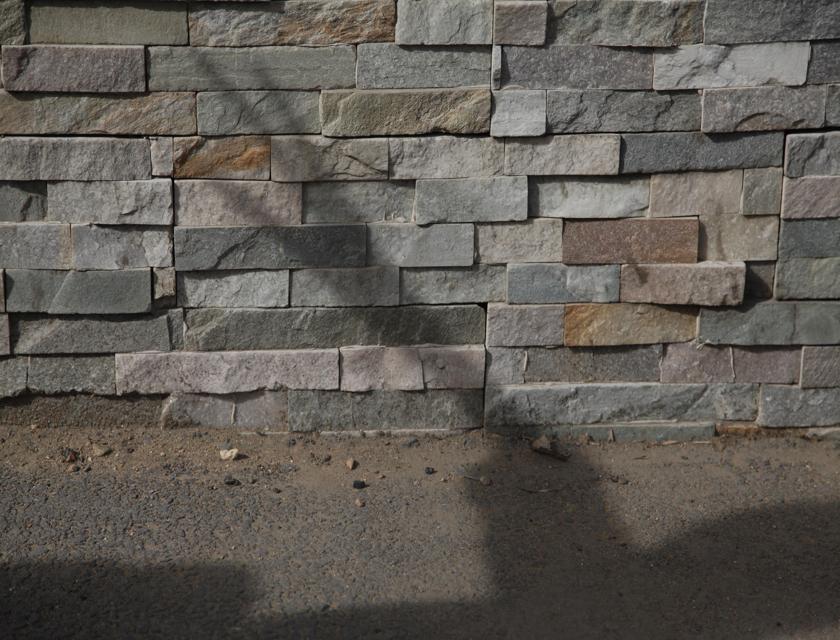}&
\includegraphics[width=0.187\linewidth,height=0.13\linewidth]{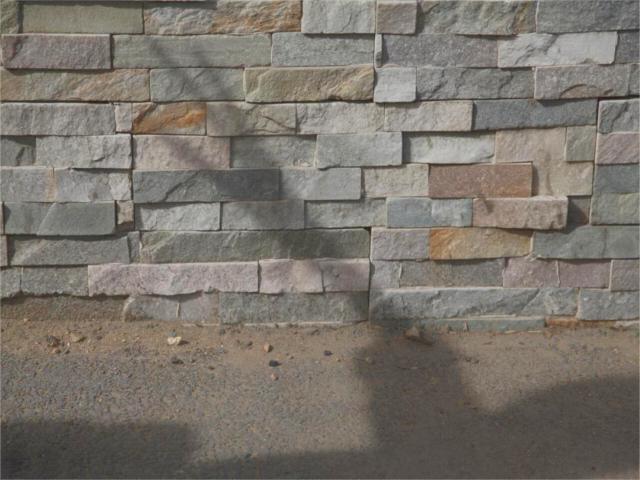}&
\includegraphics[width=0.187\linewidth,height=0.13\linewidth]{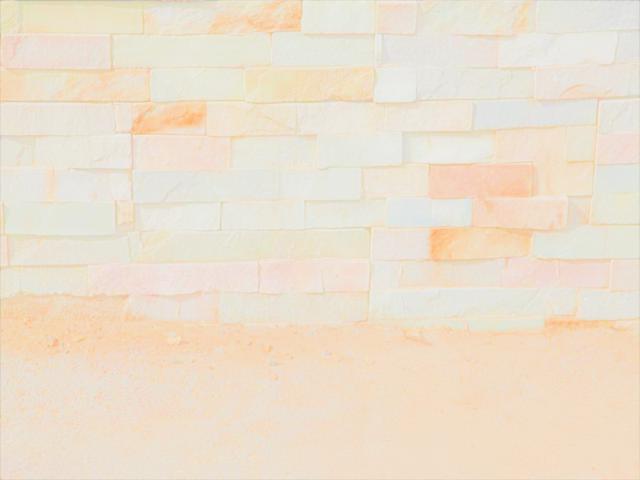}&
\includegraphics[width=0.187\linewidth,height=0.13\linewidth]{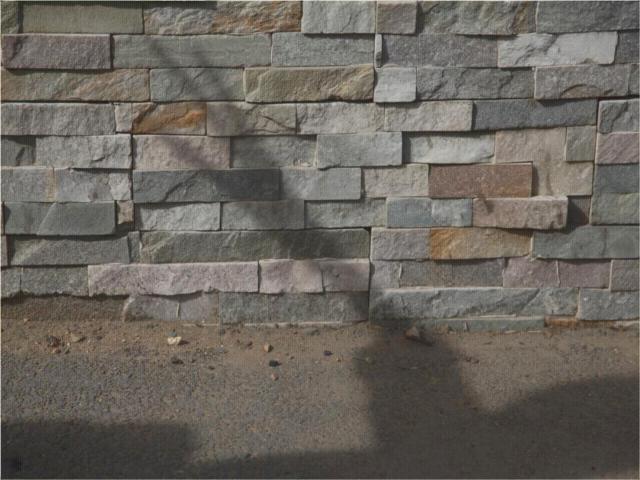}& 
\includegraphics[width=0.187\linewidth,height=0.13\linewidth]{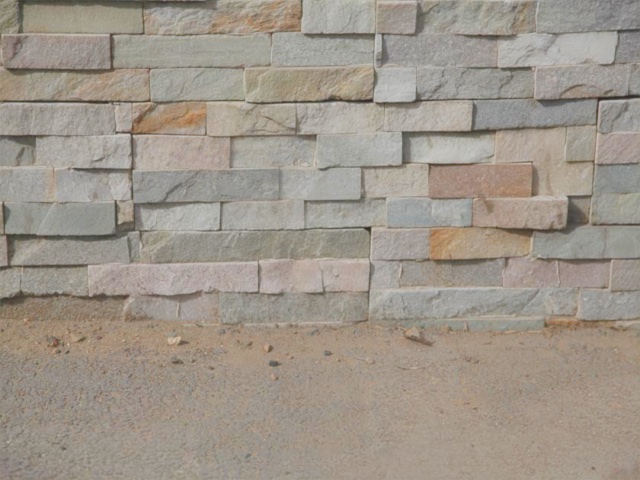}& 
\\
\includegraphics[width=0.187\linewidth,height=0.13\linewidth]{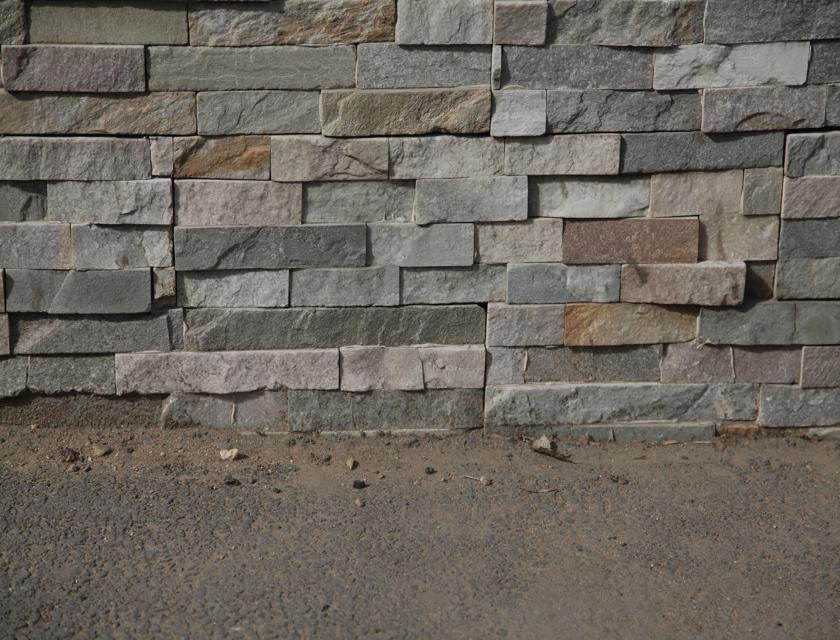}&
\includegraphics[width=0.187\linewidth,height=0.13\linewidth]{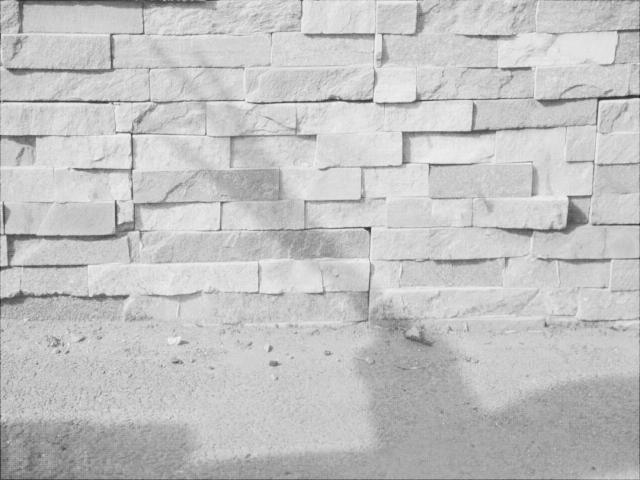}&
\includegraphics[width=0.187\linewidth,height=0.13\linewidth]{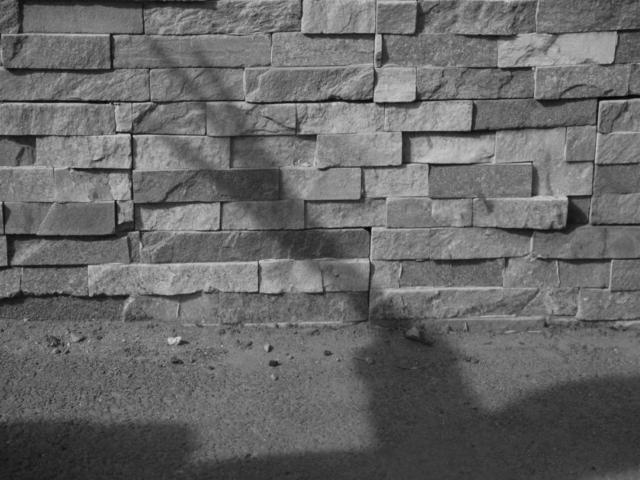}&
\includegraphics[width=0.187\linewidth,height=0.13\linewidth]{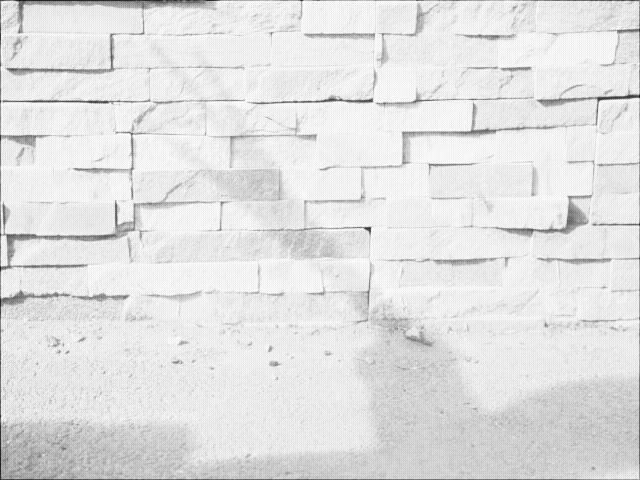}& 
\includegraphics[width=0.187\linewidth,height=0.13\linewidth]{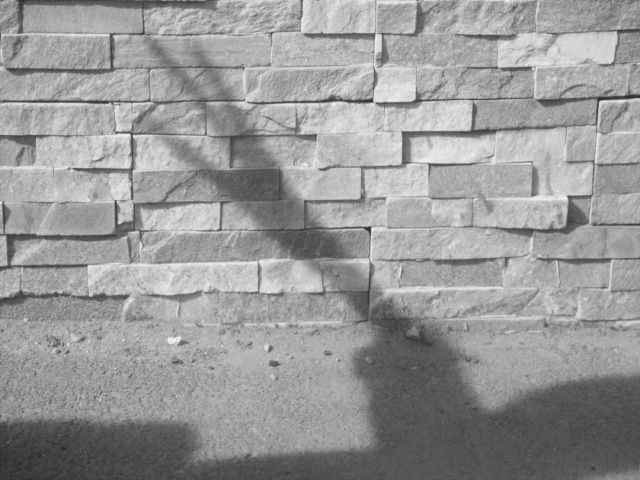}& 
\\
\fontsize{8pt}{\baselineskip}\selectfont Input $\&$ GT &
\fontsize{8pt}{\baselineskip}\selectfont  $\mathcal{L}_{\text{fid}}$& 
\fontsize{8pt}{\baselineskip}\selectfont  $\mathcal{L}_{\text{fid}}+\mathcal{L}_{\text{ill}}$& 
\fontsize{8pt}{\baselineskip}\selectfont  $\mathcal{L}_{\text{fid}}+\mathcal{L}_{\text{ref}}$& 
\fontsize{8pt}{\baselineskip}\selectfont  $\mathcal{L}_{\text{de}}$& 

\\
\end{tabular}
\end{center}
\caption{\rep{Visual comparisons of shadow decomposition of our method with different regularizations.}}
\label{fig:ablation_decomp}

\end{figure}

\rep{\figref{fig:ablation_decomp} shows visual comparisons of various regularizations. 
Shadows persist in both \rep{$\mathbf{R}_{\text{s}}$ and $\mathbf{L}_{\text{s}}$} when relying solely on $\mathcal{L}_{\text{fid}}$. 
Despite the separation of $\mathbf{R}_{\text{s}}$ and $\mathbf{L}_{\text{s}}$ when $\mathcal{L}_{\text{ill}}$ is added to $\mathcal{L}_{\text{fid}}$, $\mathbf{R}_{\text{s}}$ shows poor color quality due to inadequate regularization.
Combining $\mathcal{L}_{\text{fid}}$ and $\mathcal{L}_{\text{ref}}$ produces similar results to those of $\mathcal{L}_{\text{fid}}$.
Finally, the $\mathcal{L}_{\text{de}}$ demonstrates the indispensability of all regularizations to the final decomposition.
}

\begin{table}[t!]
\begin{center}
\footnotesize
\renewcommand\tabcolsep{1.2pt}
\renewcommand\arraystretch{0.7}
    \begin{tabular}{l|ccc|cc|cc}
    \toprule
 \multirow{2}{*}{}      & \multicolumn{3}{c|}{\rep{Type of condition}} & \multicolumn{2}{c|}{\rep{Range of denoising}} & \multicolumn{2}{c}{\textit{S}}  \\   
 & [$\mathbf{L}_{\text{s}}$] &[$\mathbf{L}_{\text{s}}\text{,} \mathbf{I}_{\text{m}} $] & [$\mathbf{L}_{\text{s}} \text{,} \mathbf{C}_t$] & Local &Global & RMSE~$\downarrow$ & PSNR~$\uparrow$ \\
 \midrule
  (a)  & \checkmark & & &&\checkmark&10.43& 31.55 \\
  (b)  & \checkmark & & &\checkmark&&9.40& 32.44 \\
(c)  & &\checkmark  & &&\checkmark&10.19& 31.92 \\
 (d)  & & \checkmark & &\checkmark&&8.57& 33.62 \\
   Ours  & & &\checkmark  &\checkmark&&{\bf 6.54}&{\bf 36.61}  \\
 \bottomrule
    \end{tabular}
    \end{center}

\caption{Ablation study of the proposed \yu{LLC} on the ISTD dataset.  
\rep{Local and Global refer to the mask regions and  whole image. For simplicity, we omit the time step $t$ in [$\cdot$].}
}
\label{tab:ablation_LLC}

\end{table}

\textbf{Local Lighting Correction Evaluation.} 
In~\tableref{tab:ablation_LLC}, we compare our LLC with four variants to evaluate its effectiveness. 
In (a) and (b), we train global and local diffusion process with only $\mathbf{L}_{s}$ as conditional input, respectively.
In (c) and (d), we add the $\mathbf{I}_{\text{m}}$ as another input condition to indicate the shadow regions.
\ke{The results demonstrate that: }
(1) 
\ke{conducting denoising process locally rather than globally can achieve better performance (see (a)/(b) or (c)/(d)),} as the spatially uniform noise distribution assumed by the global diffusion may not handle differences of intensity distributions between shadow and non-shadow regions;
(2) the comparisons of Ours to (b) and (d) prove that 
\ke{the lighting prior from non-shadow regions is crucial for shadow region illumination correction, as it uses global lighting information to efficiently constrain the  sampling space in diffusion process;}
(3) 
\ke{explicitly using the lighting prior from non-shadow regions as a \rep{time-embedded} condition \rynn{provides} further significant performance improvement (see (d) and Ours).}

\textbf{Illumination-guided texture restoration Evaluation.}
{We further evaluate our IGTR module with \yur{six} variants in~\tableref{tab:ablation_igtr}.
First, we directly composing $\mathbf{R}_{\text{s}}$ and $\hat{\mathbf{L}}_{\text{s}}$ via element-wise multiplication, referred to as ``$\mathbf{R}_{\text{s}} \times \hat{\mathbf{L}}_{\text{s}}$''. 
Second, we exclude IGTR and concatenate $\mathbf{R}_{\text{s}}$ and $\hat{\mathbf{L}}_{\text{s}}$, or their corresponding features, resulting in two baselines: ``Cat (i)'' and ``Cat (f)''. 
Next, we substitute IGTR with the standard self-attention~\cite{vaswani2017attention}, labeled as ``SA''. 
Finally, we remove the local and non-local lighting-to-texture correspondence modeling in the IGTR, separately, yielding  ``IGTR (L)'' and ``IGTR (G)''. 
} 
~\tableref{tab:ablation_igtr} indicates that omitting IGTR compromises shadow removal efficacy, with our IGTR outperforming the standard self-attention. It also demonstrates that both local and non-local lighting-to-texture correspondence modeling aids in restoring shadow region textures, and their combination optimizes results. \rep{Further, comparing Cat(i) and Cat(f), SA and/or IGTR(G) with IGTR(L) highlights the superior utility of local-based features over global-based ones.}

\begin{table}
\begin{center}
\footnotesize
    \renewcommand\tabcolsep{1.2pt}
    \renewcommand\arraystretch{0.5}
        \begin{tabular}{c|c|c|c|c|c|c|c}
        \toprule
        \textit{S} &$\mathbf{R}_{\text{s}} \times \hat{\mathbf{L}}_{\text{s}}$ &Cat (i)&Cat (f)&SA&IGTR (G)&IGTR (L)&Ours \\
         \midrule
        RMSE~$\downarrow$ & 7.37 & 7.30 & 7.20& 7.32& 7.27 & 7.00 &{\bf 6.54} \\
         \bottomrule
        \end{tabular}
    \end{center}

\caption{Ablation study of the proposed IGTR on the ISTD dataset. Cat (i) and (f) mean that the $\mathbf{R}_{\text{s}}$ and $\hat{\mathbf{L}}_{\text{s}}$ are concatenated at the input  and feature-level. SA denotes the standard self-attention. IGTR (G) and (L) refer to the non-local and local lighting-to-texture corresponding. 
}
\label{tab:ablation_igtr}

\end{table}

\begin{table}[t!]
\begin{center}
\footnotesize
		\renewcommand\tabcolsep{2.5pt}
		\renewcommand\arraystretch{0.5}
            \begin{tabular}{l|cc|cc|cc}
            \toprule
         \multirow{2}{*}{Datasets} & \multicolumn{2}{c|}{Bounding boxes}& \multicolumn{2}{c|}{Dilation masks} & \multicolumn{2}{c}{GT masks} \\   
         & RMSE$\downarrow$ & PSNR$\uparrow$ & RMSE$\downarrow$ & PSNR$\uparrow$ &RMSE$\downarrow$ & PSNR$\uparrow$ \\
         \midrule
         SRD  &5.69&36.39&5.49&36.51&5.40&36.79 \\
         ISTD  &6.70&35.97&6.54&36.61&5.98&37.65 \\
         ISTD+  &5.82&37.77&5.69&38.04&5.63&38.34 \\
         \bottomrule
            \end{tabular}
    \end{center}

\caption{Our method is robust to different types of shadow {annotations. \yu{Metrics are evaluated in the shadow regions.}}}
\label{tab:ablation_mask}

\end{table}

\begin{table}[t!]
\begin{center}

\footnotesize
\renewcommand\tabcolsep{2pt}
\renewcommand\arraystretch{0.5}
\begin{tabular}{l|c|c|c|c|c}
\toprule
 Methods   & G2R  & AEF  & SP+M+I   & SGNet  & Ours\\
 \midrule
 Params. (MB) &27.75 &143.01 &195.6 &6.2 &171.87 \\
 Time (s) &0.32&0.14 &0.15 &0.25 &1.70 \\
 \bottomrule

\end{tabular}
\caption{Comparisons of parameters and inference time.}
\label{tab:time_comp}

\end{center}
\end{table}

\textbf{Robustness to Shadow Annotations.} 
 In \tableref{tab:ablation_mask},  we compare the shadow removal results using our method with bounding boxes, dilated~\cite{fu2021auto}, and ground truth masks. 
 Benefiting from the proposed shadow-aware decomposition that constrains shadows to illumination layer, our method is robust to the accuracy of shadow masks. 
Notably, our method  \yu{using bounding box} outperforms the BMNet \rep{using shadow mask} with 11.8\% RMSE reduction in the shadow regions \yu{on the ISTD dataset}. 

\section{Conclusion}
\label{sec:conclusion}
In this paper, we have proposed a novel method for shadow removal, which
includes a shadow-aware decomposition network to derive the shadow reflectance and illumination layers. A novel bilateral correction network is proposed with a novel LLC module and a novel IGTR module, to re-cast the \yu{degraded} lighting and restore the degraded textures in shadow regions conditionally.
We have also annotated the shadow masks for the SRD benchmark, for a fair evaluation with existing shadow removal methods.
We conduct extensive experiments on three shadow removal benchmarks, to demonstrate the superior performance of our method. %

Despite its efficacy, our method has limitations, including slightly larger parameters and  extended inference times due to the use of diffusion, as shown in Tab~\ref{tab:time_comp}. For instance, our method requires 1.7 seconds to process a 640 $\times$ 480 image. 

\section{Acknowledgments}
This project is in part supported by two SRG grants from the City University of Hong Kong (No.: 7005674 and 7005843).

\bibliography{aaai24}

\begin{thebibliography}{56}
\providecommand{\natexlab}[1]{#1}

\bibitem[{Baslamisli, Le, and Gevers(2018)}]{baslamisli2018cnn}
Baslamisli, A.~S.; Le, H.-A.; and Gevers, T. 2018.
\newblock CNN based learning using reflection and retinex models for intrinsic image decomposition.
\newblock In \emph{CVPR}.

\bibitem[{Brown and Tsoi(2006)}]{brown2006geometric}
Brown, M.~S.; and Tsoi, Y.-C. 2006.
\newblock Geometric and shading correction for images of printed materials using boundary.
\newblock \emph{IEEE TIP}.

\bibitem[{Chen et~al.(2021)Chen, Long, Zhang, and Xiao}]{chen2021canet}
Chen, Z.; Long, C.; Zhang, L.; and Xiao, C. 2021.
\newblock CANet: A Context-Aware Network for Shadow Removal.
\newblock In \emph{ICCV}.

\bibitem[{Cun, Pun, and Shi(2020)}]{cun2020towards}
Cun, X.; Pun, C.-M.; and Shi, C. 2020.
\newblock Towards ghost-free shadow removal via dual hierarchical aggregation network and shadow matting GAN.
\newblock In \emph{AAAI}.

\bibitem[{Dhariwal and Nichol(2021)}]{dhariwal2021diffusion}
Dhariwal, P.; and Nichol, A. 2021.
\newblock Diffusion models beat gans on image synthesis.
\newblock In \emph{NeurIPS}.

\bibitem[{Ding et~al.(2019)Ding, Long, Zhang, and Xiao}]{ding2019argan}
Ding, B.; Long, C.; Zhang, L.; and Xiao, C. 2019.
\newblock Argan: Attentive recurrent generative adversarial network for shadow detection and removal.
\newblock In \emph{ICCV}.

\bibitem[{Drew, Finlayson, and Hordley(2003)}]{drew2003recovery}
Drew, M.~S.; Finlayson, G.~D.; and Hordley, S.~D. 2003.
\newblock Recovery of chromaticity image free from shadows via illumination invariance.
\newblock In \emph{ICCVW}.

\bibitem[{Finlayson and Drew(2001)}]{finlayson2001}
Finlayson, G.~D.; and Drew, M.~S. 2001.
\newblock 4-sensor camera calibration for image representation invariant to shading, shadows, lighting, and specularities.
\newblock In \emph{ICCV}.

\bibitem[{Finlayson, Drew, and Lu(2009)}]{finlayson2009entropy}
Finlayson, G.~D.; Drew, M.~S.; and Lu, C. 2009.
\newblock Entropy minimization for shadow removal.
\newblock \emph{IJCV}.

\bibitem[{Finlayson, Hordley, and Drew(2002)}]{finlayson2002removing}
Finlayson, G.~D.; Hordley, S.~D.; and Drew, M.~S. 2002.
\newblock Removing shadows from images using retinex.
\newblock In \emph{CI}.

\bibitem[{Finlayson et~al.(2005)Finlayson, Hordley, Lu, and Drew}]{finlayson2005removal}
Finlayson, G.~D.; Hordley, S.~D.; Lu, C.; and Drew, M.~S. 2005.
\newblock On the removal of shadows from images.
\newblock \emph{IEEE TPAMI}.

\bibitem[{Fu et~al.(2021)Fu, Zhou, Guo, Juefei-Xu, Yu, Feng, Liu, and Wang}]{fu2021auto}
Fu, L.; Zhou, C.; Guo, Q.; Juefei-Xu, F.; Yu, H.; Feng, W.; Liu, Y.; and Wang, S. 2021.
\newblock Auto-exposure fusion for single-image shadow removal.
\newblock In \emph{CVPR}.

\bibitem[{Gryka, Terry, and Brostow(2015)}]{gryka2015learning}
Gryka, M.; Terry, M.; and Brostow, G.~J. 2015.
\newblock Learning to remove soft shadows.
\newblock \emph{ACM TOG}.

\bibitem[{Guo, Dai, and Hoiem(2012)}]{guo2012paired}
Guo, R.; Dai, Q.; and Hoiem, D. 2012.
\newblock Paired regions for shadow detection and removal.
\newblock \emph{IEEE TPAMI}.

\bibitem[{He et~al.(2021)He, Xing, Zhang, and Chen}]{he2021unsupervised}
He, Y.; Xing, Y.; Zhang, T.; and Chen, Q. 2021.
\newblock Unsupervised Portrait Shadow Removal via Generative Priors.
\newblock In \emph{ACM MM}.

\bibitem[{Ho, Jain, and Abbeel(2020)}]{ho2020denoising}
Ho, J.; Jain, A.; and Abbeel, P. 2020.
\newblock Denoising diffusion probabilistic models.
\newblock In \emph{NeurIPS}.

\bibitem[{Hu et~al.(2019{\natexlab{a}})Hu, Fu, Zhu, Qin, and Heng}]{hu2019direction}
Hu, X.; Fu, C.-W.; Zhu, L.; Qin, J.; and Heng, P.-A. 2019{\natexlab{a}}.
\newblock Direction-aware spatial context features for shadow detection and removal.
\newblock \emph{IEEE TPAMI}.

\bibitem[{Hu et~al.(2019{\natexlab{b}})Hu, Jiang, Fu, and Heng}]{hu2019maskshadowgan}
Hu, X.; Jiang, Y.; Fu, C.-W.; and Heng, P.-A. 2019{\natexlab{b}}.
\newblock Mask-ShadowGAN: Learning to remove shadows from unpaired data.
\newblock In \emph{ICCV}.

\bibitem[{Jin, Sharma, and Tan(2021)}]{jin2021dc}
Jin, Y.; Sharma, A.; and Tan, R.~T. 2021.
\newblock DC-ShadowNet: Single-Image Hard and Soft Shadow Removal Using Unsupervised Domain-Classifier Guided Network.
\newblock In \emph{ICCV}.

\bibitem[{Johnson, Alahi, and Fei-Fei(2016)}]{johnson2016perceptual}
Johnson, J.; Alahi, A.; and Fei-Fei, L. 2016.
\newblock Perceptual losses for real-time style transfer and super-resolution.
\newblock In \emph{ECCV}.

\bibitem[{Khan et~al.(2015)Khan, Bennamoun, Sohel, and Togneri}]{khan2015automatic}
Khan, S.~H.; Bennamoun, M.; Sohel, F.; and Togneri, R. 2015.
\newblock Automatic shadow detection and removal from a single image.
\newblock \emph{IEEE TPAMI}.

\bibitem[{Kingma and Ba(2015)}]{diederik2015adam}
Kingma, D.~P.; and Ba, J. 2015.
\newblock Adam: {A} Method for Stochastic Optimization.
\newblock In \emph{ICLR}.

\bibitem[{Land(1977)}]{land1977retinex}
Land, E.~H. 1977.
\newblock The retinex theory of color vision.
\newblock \emph{Scientific American}.

\bibitem[{Le and Samaras(2019)}]{le2019shadow}
Le, H.; and Samaras, D. 2019.
\newblock Shadow removal via shadow image decomposition.
\newblock In \emph{ICCV}.

\bibitem[{Le and Samaras(2020)}]{le2020from}
Le, H.; and Samaras, D. 2020.
\newblock From shadow segmentation to shadow removal.
\newblock In \emph{ECCV}.

\bibitem[{Le and Samaras(2021)}]{le2021physics}
Le, H.; and Samaras, D. 2021.
\newblock Physics-based shadow image decomposition for shadow removal.
\newblock \emph{IEEE TPAMI}.

\bibitem[{Liu et~al.(2023{\natexlab{a}})Liu, Liu, Kong, Xu, Zhang, Yin, Hancke, and Lau}]{liu2023referring}
Liu, F.; Liu, Y.; Kong, Y.; Xu, K.; Zhang, L.; Yin, B.; Hancke, G.; and Lau, R. 2023{\natexlab{a}}.
\newblock Referring image segmentation using text supervision.
\newblock In \emph{ICCV}, 22124--22134.

\bibitem[{Liu et~al.(2023{\natexlab{b}})Liu, Guo, Fu, Ke, Xu, Feng, Tsang, and Lau}]{liu2023structure}
Liu, Y.; Guo, Q.; Fu, L.; Ke, Z.; Xu, K.; Feng, W.; Tsang, I.~W.; and Lau, R.~W. 2023{\natexlab{b}}.
\newblock Structure-Informed Shadow Removal Networks.
\newblock \emph{IEEE TIP}.

\bibitem[{Liu et~al.(2021{\natexlab{a}})Liu, Yin, Mi, Pu, and Wang}]{liu2021shadow}
Liu, Z.; Yin, H.; Mi, Y.; Pu, M.; and Wang, S. 2021{\natexlab{a}}.
\newblock Shadow removal by a lightness-guided network with training on unpaired data.
\newblock \emph{IEEE TIP}.

\bibitem[{Liu et~al.(2021{\natexlab{b}})Liu, Yin, Wu, Wu, Mi, and Wang}]{liu2021from}
Liu, Z.; Yin, H.; Wu, X.; Wu, Z.; Mi, Y.; and Wang, S. 2021{\natexlab{b}}.
\newblock From Shadow Generation to Shadow Removal.
\newblock In \emph{CVPR}.

\bibitem[{Mei et~al.(2021)Mei, Ji, Wei, Yang, Wei, and Fan}]{mei2021camouflaged}
Mei, H.; Ji, G.-P.; Wei, Z.; Yang, X.; Wei, X.; and Fan, D.-P. 2021.
\newblock Camouflaged object segmentation with distraction mining.
\newblock In \emph{CVPR}.

\bibitem[{Meka et~al.(2021)Meka, Shafiei, Zollh{\"o}fer, Richardt, and Theobalt}]{meka2021real}
Meka, A.; Shafiei, M.; Zollh{\"o}fer, M.; Richardt, C.; and Theobalt, C. 2021.
\newblock Real-time global illumination decomposition of videos.
\newblock \emph{ACM TOG}.

\bibitem[{Qu et~al.(2017)Qu, Tian, He, Tang, and Lau}]{qu2017deshadownet}
Qu, L.; Tian, J.; He, S.; Tang, Y.; and Lau, R.~W. 2017.
\newblock Deshadownet: A multi-context embedding deep network for shadow removal.
\newblock In \emph{CVPR}.

\bibitem[{Rombach et~al.(2022)Rombach, Blattmann, Lorenz, Esser, and Ommer}]{rombach2022high}
Rombach, R.; Blattmann, A.; Lorenz, D.; Esser, P.; and Ommer, B. 2022.
\newblock High-resolution image synthesis with latent diffusion models.
\newblock In \emph{CVPR}.

\bibitem[{Saharia et~al.(2022{\natexlab{a}})Saharia, Chan, Chang, Lee, Ho, Salimans, Fleet, and Norouzi}]{saharia2022palette}
Saharia, C.; Chan, W.; Chang, H.; Lee, C.; Ho, J.; Salimans, T.; Fleet, D.; and Norouzi, M. 2022{\natexlab{a}}.
\newblock Palette: Image-to-image diffusion models.
\newblock In \emph{SIGGRAPH}.

\bibitem[{Saharia et~al.(2022{\natexlab{b}})Saharia, Ho, Chan, Salimans, Fleet, and Norouzi}]{saharia2022image}
Saharia, C.; Ho, J.; Chan, W.; Salimans, T.; Fleet, D.~J.; and Norouzi, M. 2022{\natexlab{b}}.
\newblock Image super-resolution via iterative refinement.
\newblock \emph{IEEE TPAMI}.

\bibitem[{Serrano et~al.(2021)Serrano, Chen, Wang, Piovar{\v{c}}i, Seidel, Didyk, and Myszkowski}]{serrano2021effect}
Serrano, A.; Chen, B.; Wang, C.; Piovar{\v{c}}i, M.; Seidel, H.-P.; Didyk, P.; and Myszkowski, K. 2021.
\newblock The effect of shape and illumination on material perception: model and applications.
\newblock \emph{ACM TOG}.

\bibitem[{Sohl-Dickstein et~al.(2015)Sohl-Dickstein, Weiss, Maheswaranathan, and Ganguli}]{sohl2015deep}
Sohl-Dickstein, J.; Weiss, E.; Maheswaranathan, N.; and Ganguli, S. 2015.
\newblock Deep unsupervised learning using nonequilibrium thermodynamics.
\newblock In \emph{ICML}.

\bibitem[{Sun et~al.(2023)Sun, Xu, Pang, Zhang, Lu, Hancke, and Lau}]{Sun_2023_ICCV}
Sun, J.; Xu, K.; Pang, Y.; Zhang, L.; Lu, H.; Hancke, G.; and Lau, R. 2023.
\newblock Adaptive Illumination Mapping for Shadow Detection in Raw Images.
\newblock In \emph{ICCV}.

\bibitem[{Vaswani et~al.(2017)Vaswani, Shazeer, Parmar, Uszkoreit, Jones, Gomez, Kaiser, and Polosukhin}]{vaswani2017attention}
Vaswani, A.; Shazeer, N.; Parmar, N.; Uszkoreit, J.; Jones, L.; Gomez, A.~N.; Kaiser, {\L}.; and Polosukhin, I. 2017.
\newblock Attention is all you need.
\newblock In \emph{NeurIPS}.

\bibitem[{Wan et~al.(2022)Wan, Yin, Wu, Wu, Liu, and Wang}]{wan2022style}
Wan, J.; Yin, H.; Wu, Z.; Wu, X.; Liu, Y.; and Wang, S. 2022.
\newblock Style-Guided Shadow Removal.
\newblock In \emph{ECCV}.

\bibitem[{Wang, Li, and Yang(2018)}]{wang2018stacked}
Wang, J.; Li, X.; and Yang, J. 2018.
\newblock Stacked conditional generative adversarial networks for jointly learning shadow detection and shadow removal.
\newblock In \emph{CVPR}.

\bibitem[{Wang et~al.(2019)Wang, Zhang, Fu, Shen, Zheng, and Jia}]{wang2019underexposed}
Wang, R.; Zhang, Q.; Fu, C.-W.; Shen, X.; Zheng, W.-S.; and Jia, J. 2019.
\newblock Underexposed photo enhancement using deep illumination estimation.
\newblock In \emph{CVPR}.

\bibitem[{Wei et~al.(2018)Wei, Wang, Yang, and Liu}]{wei2018deep}
Wei, C.; Wang, W.; Yang, W.; and Liu, J. 2018.
\newblock Deep retinex decomposition for low-light enhancement.
\newblock In \emph{BMVC}.

\bibitem[{Wu et~al.(2007)Wu, Tang, Brown, and Shum}]{wu2007natural}
Wu, T.-P.; Tang, C.-K.; Brown, M.~S.; and Shum, H.-Y. 2007.
\newblock Natural shadow matting.
\newblock \emph{ACM TOG}.

\bibitem[{Xia et~al.(2022)Xia, Pan, Song, Li, and Huang}]{xia2022vision}
Xia, Z.; Pan, X.; Song, S.; Li, L.~E.; and Huang, G. 2022.
\newblock Vision transformer with deformable attention.
\newblock In \emph{CVPR}.

\bibitem[{Xu, Hancke, and Lau(2023)}]{Xu_2023_ICCV}
Xu, K.; Hancke, G.~P.; and Lau, R.~W. 2023.
\newblock Learning Image Harmonization in the Linear Color Space.
\newblock In \emph{ICCV}.

\bibitem[{Yang, Tan, and Ahuja(2012)}]{yang2012shadow}
Yang, Q.; Tan, K.-H.; and Ahuja, N. 2012.
\newblock Shadow removal using bilateral filtering.
\newblock \emph{IEEE TIP}.

\bibitem[{Zhang and Agrawala(2023)}]{zhang2023adding}
Zhang, L.; and Agrawala, M. 2023.
\newblock Adding Conditional Control to Text-to-Image Diffusion Models.
\newblock \emph{arXiv preprint arXiv:2302.05543}.

\bibitem[{Zhang et~al.(2020{\natexlab{a}})Zhang, Long, Zhang, and Xiao}]{zhang2020ris}
Zhang, L.; Long, C.; Zhang, X.; and Xiao, C. 2020{\natexlab{a}}.
\newblock Ris-gan: Explore residual and illumination with generative adversarial networks for shadow removal.
\newblock In \emph{AAAI}.

\bibitem[{Zhang, Zhang, and Xiao(2015)}]{zhang2015shadow}
Zhang, L.; Zhang, Q.; and Xiao, C. 2015.
\newblock Shadow remover: Image shadow removal based on illumination recovering optimization.
\newblock \emph{IEEE TIP}.

\bibitem[{Zhang et~al.(2020{\natexlab{b}})Zhang, Zhao, Li, and Wang}]{zhang2020shadow}
Zhang, M.; Zhao, W.; Li, X.; and Wang, D. 2020{\natexlab{b}}.
\newblock Shadow detection of moving objects in traffic monitoring video.
\newblock In \emph{ITAIC}.

\bibitem[{Zhang et~al.(2021)Zhang, Guo, Ma, Liu, and Zhang}]{zhang2021beyond}
Zhang, Y.; Guo, X.; Ma, J.; Liu, W.; and Zhang, J. 2021.
\newblock Beyond brightening low-light images.
\newblock \emph{IJCV}.

\bibitem[{Zhu et~al.(2021)Zhu, Xu, Ke, and Lau}]{zhu2021mitigating}
Zhu, L.; Xu, K.; Ke, Z.; and Lau, R.~W. 2021.
\newblock Mitigating Intensity Bias in Shadow Detection via Feature Decomposition and Reweighting.
\newblock In \emph{ICCV}.

\bibitem[{Zhu et~al.(2022{\natexlab{a}})Zhu, Huang, Fu, Zhao, Sun, and Zha}]{zhu2022bijective}
Zhu, Y.; Huang, J.; Fu, X.; Zhao, F.; Sun, Q.; and Zha, Z.-J. 2022{\natexlab{a}}.
\newblock Bijective Mapping Network for Shadow Removal.
\newblock In \emph{CVPR}.

\bibitem[{Zhu et~al.(2022{\natexlab{b}})Zhu, Xiao, Fang, Fu, Xiong, and Zha}]{zhu2022efficient}
Zhu, Y.; Xiao, Z.; Fang, Y.; Fu, X.; Xiong, Z.; and Zha, Z.-J. 2022{\natexlab{b}}.
\newblock Efficient Model-Driven Network for Shadow Removal.
\newblock In \emph{AAAI}.

\end{thebibliography}

\end{document}